\documentclass[a4paper,UKenglish,cleveref, autoref, thm-restate]{lipics-v2021}
\usepackage{booktabs}
\usepackage{subcaption}
\usepackage{graphicx}
\usepackage[autostyle]{csquotes}
\usepackage{comment}
\MakeOuterQuote{"}
\usepackage{xcolor}
\definecolor{lightpink}{RGB}{215, 115, 135} 
\definecolor{lightgreen}{RGB}{85, 165, 85}  

\title{LMM Modality Transfer: A Pre-requisite for Autonomous GIS Agents} 

\author{Ivan Majic\footnote{corresponding author}}{Graz University of Technology, Austria \and University of Vienna, Austria}{majic@tugraz.at}{https://orcid.org/0000-0002-0834-3791}{}
\author{Zexian Huang}{Graz University of Technology, Austria} {zexian.huang@tugraz.at}{https://orcid.org/0000-0003-0270-1604}{}
\author{Franziska Hübl}{Graz University of Technology, Austria} {franziska.huebl@tugraz.at}{https://orcid.org/0000-0002-3548-2455}{}
\author{Krzysztof Janowicz}{University of Vienna, Austria}{krzysztof.janowicz@univie.ac.at}{https://orcid.org/0009-0003-1968-887X}{}
\author{Meilin Shi}{University of Liverpool, UK}{meilin.shi@liverpool.ac.uk}{https://orcid.org/0000-0001-6039-7810}{}
\author{Mina Karimi}{University of Vienna, Austria}{mina.karimi@univie.ac.at}{https://orcid.org/0000-0003-2521-8164}{}
\author{Zilong Liu}{University of Vienna, Austria}{zilong.liu@univie.ac.at}{https://orcid.org/0000-0002-7699-3366}{}
\author{Alexandra Fortacz-Lazan}{University of Vienna, Austria}{alexandra.fortacz@univie.ac.at}{https://orcid.org/0009-0007-8370-1572}{}

\authorrunning{Majic et al.}

\Copyright{Ivan Majic, Zexian Huang, Franziska Huebl, Krzysztof Janowicz, Meilin Shi, Mina Karimi, Zilong Liu, and Alexandra Fortacz} 

\supplementdetails [subcategory={Source Code}]{Software}{https://github.com/Geoinfo-TUGraz/COSIT2026_LMM_modality_transfer}



\nolinenumbers 

\EventEditors{Sabine Timpf, Gabriele Filomena, Armand Kapaj, Rui Zhu, Nicholas Giudice, and Ed Manley}
\EventNoEds{6}
\EventLongTitle{17th International Conference on Spatial Information Theory (COSIT 2026)}
\EventShortTitle{COSIT 2026}
\EventAcronym{COSIT}
\EventYear{2026}
\EventDate{September 22--25, 2026}
\EventLocation{York, UK}
\EventLogo{}
\SeriesVolume{393}
\ArticleNo{14}

\begin{document}

\maketitle

\begin{abstract}
AI models are becoming increasingly adept at understanding and processing spatial information, thereby facilitating agentic problem-solving in spatial tasks and workflows. However, most of the research on their spatial capabilities (e.g., spatial reasoning) has focused on the textual modality as input and output. This contrasts with the human approach to GIS workflows, where text and visual modalities are often used together, interchangeably, and in a complementary manner. Thus, to truly achieve an automated GIS analysis pipeline or carry out human-designed GIS workflows, AI models --- Large Multimodal Models (LMMs) in particular --- need to be able to seamlessly transition between image- and text-based modalities that are traditionally used in such workflows. We present a modality transfer task that (1) asks an LMM to first describe an input image of colored squares in a regular grid, and (2) asks a new LMM instance to re-generate an image of the original spatial scene using the textual description output by the former model. This task quantifies the ability of LMMs to transfer spatial information between image and text modalities. Ultimately, by examining the modality transfer capability of LMMs through the lens of spatial information theory, this work highlights a critical bottleneck: achieving strong and robust geospatial understanding in LMMs requires rigorous, multi-modal alignment. Our results indicate that recent LMMs (here from OpenAI) still struggle with modality transfer, when tasked with re-generating an image of a simple spatial grid of color squares.
\end{abstract}

\section{Introduction}
\label{sec:Introduction}
Typical GIS workflows inherently use both text and image modalities interchangeably, depending on the situation and the communication needs. For example, two human GIS analysts who work side by side on a joint project may exchange textual information like feature attributes, but also discuss the semantic meaning of certain colored polygons by asking questions like \emph{``what does this elongated purple shape represent''}. If one or both of these GIS analysts were to be replaced by autonomous GIS agents, we argue that the same (human-expert) level of understanding and seamless transition between both modalities of geographic information is a pre-requisite for smooth, reliable, and deterministically repeatable operation \cite{scheider2015talk}.

Development of such intelligent GIS agents is a continuation of the long-standing desire and need to automate GIS workflows. Tasks such as information retrieval, data preparation, and data cleaning have always been resource-hungry. This is why many computational workflows and task-specific GIS automations have been developed as plugins to GIS environments such as ArcGIS or QGIS \cite{Majic2019b}, or as standalone software tools~\cite{Majic2019,Majic2021}. With the advent of generative AI and foundation models that are capable of addressing a variety of downstream tasks without specific pre-training, the aspirations of autonomous GIS agents have also risen \cite{Li2025}.
An intelligent GIS agent that can understand the data it is working with and autonomously adjust the workflows to the specific situation would be of great value. Hence, such a \textit{GeoMachina} has been declared one of the potential common moonshots for the field of GeoAI \cite{Janowicz2020}. While there is substantial progress \cite{Li2025,akinboyewa2025gis}, some fundamental considerations regarding the multi-modality of spatial information remain unaddressed. 

Thus, we bring forth an essential question regarding the underlying foundation model infrastructure and its readiness to fully autonomously handle inherently multi-modal GIS workflows: 
\emph{"how well can large multimodal models (LMMs) understand each other and can they transfer spatial information between the two modalities (image and text) without losing information and quality?"}
To this end, we have designed a modality transfer task that takes images of (spatially arranged) grids of colored cells (the likes of which can be found in typical land cover maps), prompts the LMM to textually describe the spatial configuration visible in the image, and then asks another model to generate an image based on this textual output. In other words, the proposed task assesses the ability of LMMs to perform \texttt{image} $\rightarrow$ \texttt{text} $\rightarrow$ \texttt{image} modality transfer. Our case study applies this task on recent-generation OpenAI models with different grid sizes and numbers of colors. \textbf{Thus, the contributions of this study are two-fold}:
\begin{itemize}
    \item We identify the lossless modality transfer as a fundamental requirement for many GIS tasks to be solved by autonomous GIS agents.
    \item We define a modality transfer task that can be used to benchmark the progress of spatial capabilities of LMMs in this context.
\end{itemize}
Thereby, our work addresses the important challenge of aligning the progress of the state-of-the-art foundational AI models with the context of GIScience by testing their ability to comprehend geographical information.

The remainder of this paper is organized as follows. In Section \ref{sec:background}, we review the literature related to our study. We propose our approach in Section \ref{sec:approach}. Then, the results are explained in Section \ref{sec:result}. We discuss our finding in Section \ref{sec:disscussion}. Finally, Section \ref{sec:conclsuion} concludes the paper.

\section{Related work}
\label{sec:background}

The rapid evolution of GeoAI has significantly advanced geographic knowledge discovery and spatial analysis. Previous foundational work has highlighted the transformative potential of spatially explicit AI techniques in addressing complex geographic problems, moving beyond traditional analysis to intelligent, agentic workflows \cite{Janowicz2020,mai2023opportunities}. While they outline various fundamental challenges in GeoAI, in particular, spatial heterogeneity, scale dependencies, and the need for spatially explicit models, the specific challenges of \emph{large multi-modal} geographic reasoning remain largely unaddressed. In contrast, single-mode spatial reasoning has received increasing attention  \cite{cohn_et_al2024,ji2025foundation,cohn2024can,gardelakosCanLargeReasoning2025}. Work has also focused on vision language models and their ability to answer spatial reasoning questions \cite{cheng2024spatialrgpt}. Recently, Xie et al. demonstrated that LLMs still lack a deeper understanding of spatial relations but rely on linguistic patterns instead \cite{xie2025evaluating}.

Traditional geographic information and human-driven GIS workflows are inherently multi-modal. Human analysts seamlessly blend visual information (e.g., maps, satellite imagery, and topological configurations) with textual information (e.g., attributes, spatial queries, and descriptive metadata). 
However, even recent GeoAI and foundation model research evaluates these modalities independently \cite{ji2025foundation}. Recent work has also shown that foundation models yield uneven success depending on which of these modalities the spatial reasoning task is presented in \cite{Majic2024}, showcasing there is a difference in spatial reasoning capabilities in different modalities. \textbf{Our work directly addresses this gap.} By introducing a framework to explicitly test the multi-modal spatial reasoning capabilities of LMMs, we study the lossless transfer of spatial information between visual and textual modalities as a critical, unaddressed prerequisite for the development of fully autonomous GeoAI agents. 

Evaluating the true representation and reasoning capabilities of AI systems requires benchmarks that isolate genuine problem-solving from mere pattern matching or data memorization. A prominent example is the Abstract and Reasoning Corpus for Artificial General Intelligence (ARC-AGI) introduced by Chollet et al. \cite{chollet2024arc}.
A guiding design philosophy of ARC-AGI is to formulate tasks that are intuitive for humans yet highly challenging for AI models, thereby exposing gaps and weaknesses in an AI's capacity to generalize abstract rules from limited examples without prior training. Our proposed modality transfer task shares this core principle but targets a different aspect. While ARC-AGI focuses on abstract logical reasoning and skill acquisition within a single visual-symbolic domain, our modality transfer task explicitly evaluates the \emph{transferability and preservation of spatial information across disparate data modalities}.

Human spatial cognition can effortlessly translate visual cues into language descriptions and vice versa. For instance, one person can look at a visual grid and verbally describe a "\textit{red square situated in the top-left corner}", and another one can accurately reproduce those visual cues solely based on the description. Our \texttt{ground-truth image} $\rightarrow$ \texttt{generated text} $\rightarrow$ \texttt{generated image} pipeline isolates this exact translation. By abstracting the task away from heavily prior GIS knowledge (such as specific coordinate reference systems or complex vector geometries) and utilizing colored spatial grids, we ensure a fair assessment. Similar to ARC-AGI, our task relies on basic spatial core priors (grids, colors, and relative positioning). However, instead of asking AI models to deduce a hidden logical rule, we test LMM's capability to accurately translate spatial reality from one modality to another. 

\section{Approach}
\label{sec:approach}
In order to assess the ability of the LMM to understand spatial information, we propose an approach that transfers the modality of a given spatial land-cover-like example from an image representation -- into a textual representation -- and back into an image representation (i.e., \texttt{image} $\rightarrow$ \texttt{generated text} $\rightarrow$ \texttt{image} modality transfer). We call this the \emph{modality transfer task}. By performing it, we can get an estimation of how well the model really understands the provided spatial configuration. We can also directly compare the final output (generated image) to the original input (image), as they are both in the same modality and, when the system performs completely correctly, should be identical. 

\begin{figure}
    \centering
    \includegraphics[width=\linewidth]{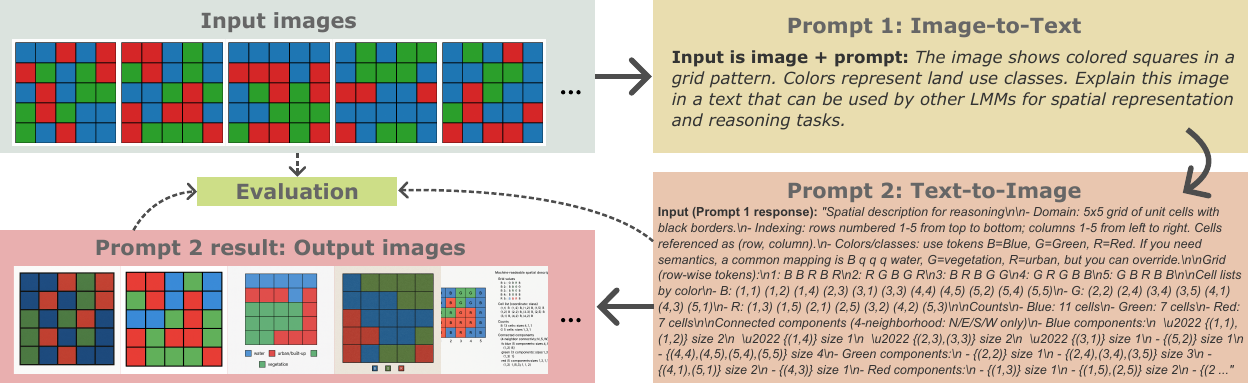}
    \caption{A diagram showing the flow of the proposed modality transfer task. Input images are generated in different configurations and batch sizes. They are then used as an input to the Prompt 1, along with a textual prompt that directs the AI how to generate a suitable textual description. The outputs of the Prompt 1 are used as inputs for the Prompt 2 where textual descriptions are used to generate new images. Finally, the generated outputs of both Prompt 1 (text) and Prompt 2 (images) are evaluated against the original input images.}
    \label{fig:diagram}
\end{figure}

The overall approach to the LMM modality transfer task is carried out as a multi-stage workflow designed to isolate, control, and quantify each step of the task. As shown in Figure~\ref{fig:diagram}, the workflow consists of the generation of input ground-truth images 
(i.e., $N \times N$  grids of colored rectangles), an LMM prompting sequence (image-to-text followed by text-to-image), the systematic extraction of generated images into discrete matrices, and a quantitative evaluation utilizing perceptual color thresholding of color matrices, text-based and image-based Levenshtein Distance \cite{levenshtein1966binary} and, Earth Mover's Distance (EMD) \cite{rubner2000earth}.

While Figure~\ref{fig:diagram} shows the logical flow of the workflow, the approach is implemented in Python programming language as a series of interactive Python notebooks. The repository containing the code, input images, and LMM prompt outputs (text and images) is published in a freely accessible repository\footnote{\url{https://github.com/Geoinfo-TUGraz/COSIT2026_LMM_modality_transfer.git}}.

\subsection{Generation of input (baseline) images}
\label{sec:image_generation}

\begin{figure}[!h]
    \centering
    \includegraphics[width=0.65\linewidth]{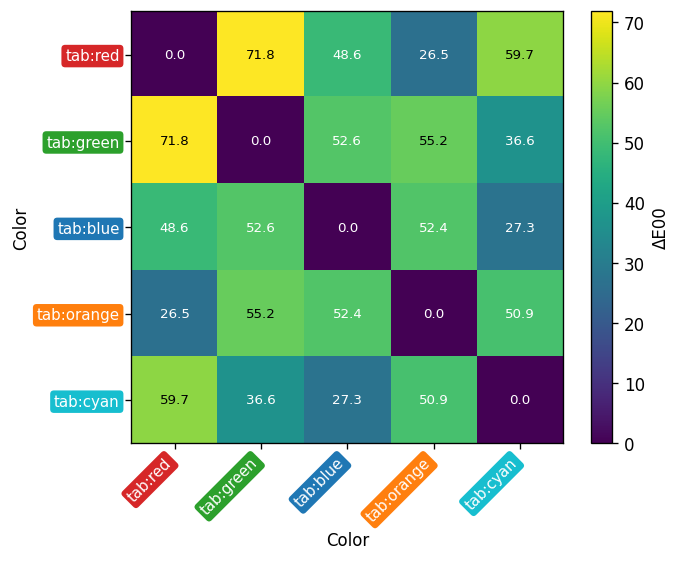}
    \caption{The five selected named colors of the Tableau Palette available in the Python Matplotlib library and the differences between them expressed as $\Delta E_{00}$.}
    \label{fig:colors}
\end{figure}

The ground-truth images consisting of $N\times N$ grids of colored rectangles are generated procedurally based on the Tableau color palette available in the Matplotlib Python library.
The five selected colors from this 10-color palette are depicted in Figure~\ref{fig:colors}, while the perceptual differences between them are expressed using the CIEDE2000 ($\Delta E_{00}$)~\cite{Luo2001} color difference formula which can express the perceived difference between two colors better than any measure based on the RGB or HSL color values~\cite{melgosaTestingCIELABbasedColordifference2000}. These distinct colors can serve as semantic labels for discrete categorical GIS data, such as land use classes, ensuring that different categories are visually distinguishable and minimizing perceptual ambiguity.

\begin{figure}[ht]
    \centering

    \begin{subfigure}[t]{0.155\textwidth}
        \centering
        \includegraphics[width=\linewidth]{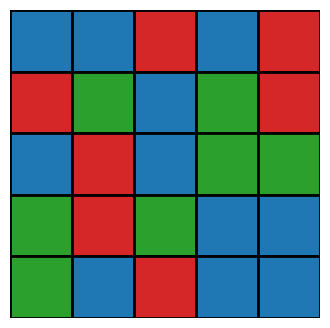}
        \caption{$5\times5\times3$}
        \label{fig:553}
    \end{subfigure}\hfill
    \begin{subfigure}[t]{0.155\textwidth}
        \centering
        \includegraphics[width=\linewidth]{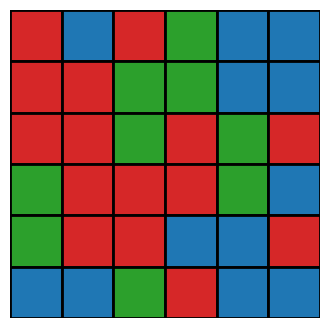}
        \caption{$6\times6\times3$}
    \end{subfigure}\hfill
    \begin{subfigure}[t]{0.155\textwidth}
        \centering
        \includegraphics[width=\linewidth]{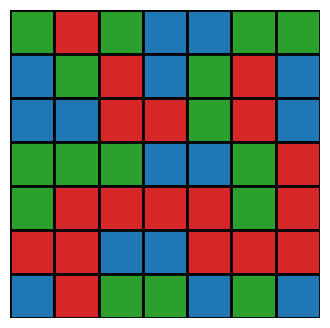}
        \caption{$7\times7\times3$}
    \end{subfigure}\hfill
    \begin{subfigure}[t]{0.155\textwidth}
        \centering
        \includegraphics[width=\linewidth]{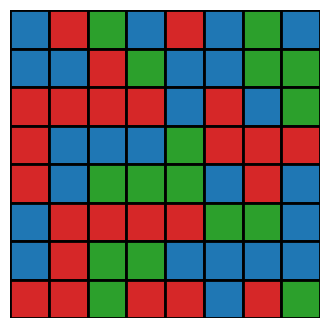}
        \caption{$8\times8\times3$}
    \end{subfigure}\hfill
    \begin{subfigure}[t]{0.155\textwidth}
        \centering
        \includegraphics[width=\linewidth]{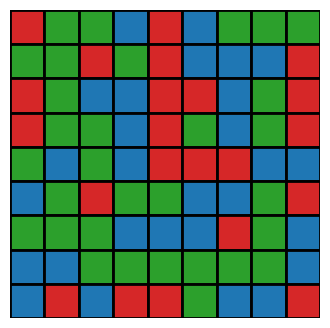}
        \caption{$9\times9\times3$}
    \end{subfigure}\hfill
    \begin{subfigure}[t]{0.155\textwidth}
        \centering
        \includegraphics[width=\linewidth]{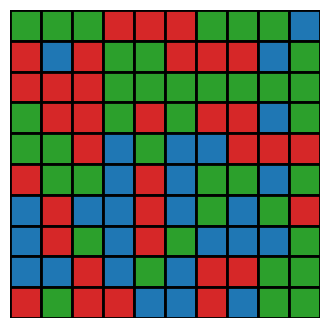}
        \caption{$10\times10\times3$}
    \end{subfigure}

    \vspace{4pt}

    \begin{subfigure}[t]{0.155\textwidth}
        \centering
        \includegraphics[width=\linewidth]{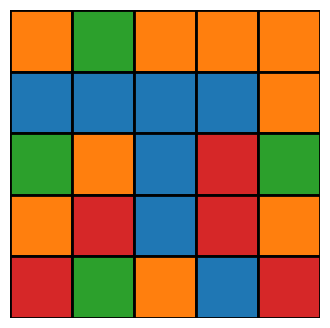}
        \caption{$5\times5\times4$}
    \end{subfigure}\hfill
    \begin{subfigure}[t]{0.155\textwidth}
        \centering
        \includegraphics[width=\linewidth]{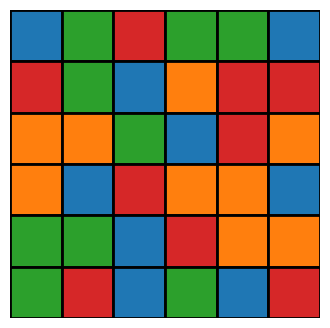}
        \caption{$6\times6\times4$}
    \end{subfigure}\hfill
    \begin{subfigure}[t]{0.155\textwidth}
        \centering
        \includegraphics[width=\linewidth]{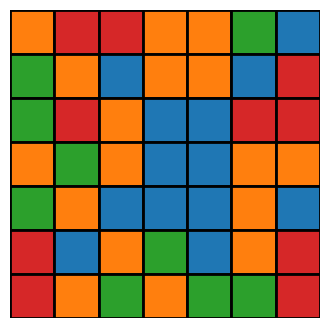}
        \caption{$7\times7\times4$}
    \end{subfigure}\hfill
    \begin{subfigure}[t]{0.155\textwidth}
        \centering
        \includegraphics[width=\linewidth]{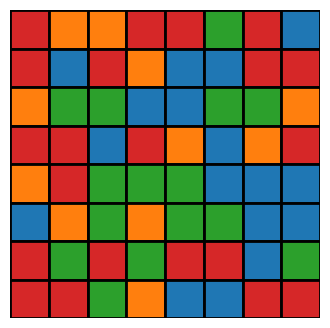}
        \caption{$8\times8\times4$}
    \end{subfigure}\hfill
    \begin{subfigure}[t]{0.155\textwidth}
        \centering
        \includegraphics[width=\linewidth]{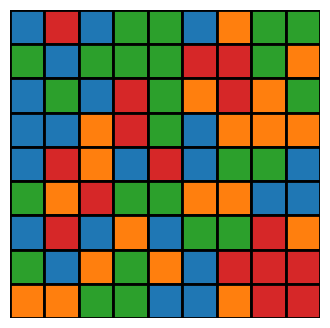}
        \caption{$9\times9\times4$}
    \end{subfigure}\hfill
    \begin{subfigure}[t]{0.155\textwidth}
        \centering
        \includegraphics[width=\linewidth]{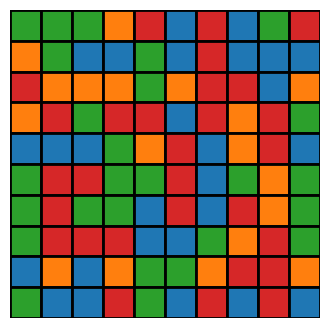}
        \caption{$10\times10\times4$}
    \end{subfigure}

    \vspace{4pt}

    \begin{subfigure}[t]{0.155\textwidth}
        \centering
        \includegraphics[width=\linewidth]{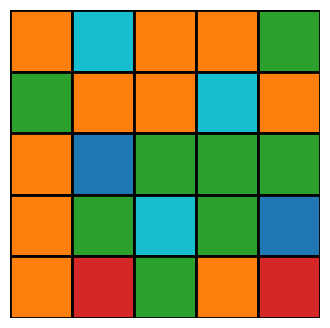}
        \caption{$5\times5\times5$}
    \end{subfigure}\hfill
    \begin{subfigure}[t]{0.155\textwidth}
        \centering
        \includegraphics[width=\linewidth]{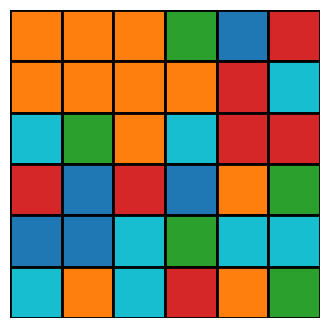}
        \caption{$6\times6\times5$}
    \end{subfigure}\hfill
    \begin{subfigure}[t]{0.155\textwidth}
        \centering
        \includegraphics[width=\linewidth]{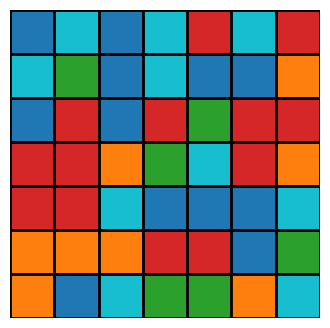}
        \caption{$7\times7\times5$}
    \end{subfigure}\hfill
    \begin{subfigure}[t]{0.155\textwidth}
        \centering
        \includegraphics[width=\linewidth]{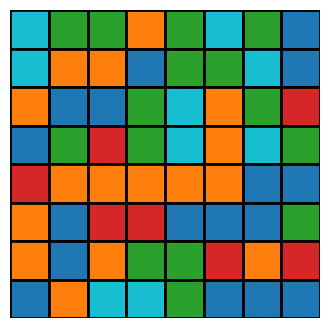}
        \caption{$8\times8\times5$}
    \end{subfigure}\hfill
    \begin{subfigure}[t]{0.155\textwidth}
        \centering
        \includegraphics[width=\linewidth]{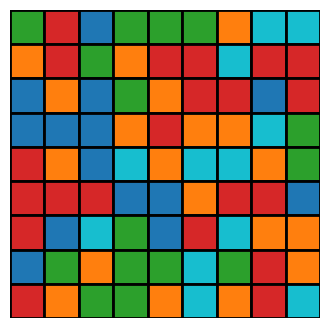}
        \caption{$9\times9\times5$}
    \end{subfigure}\hfill
    \begin{subfigure}[t]{0.155\textwidth}
        \centering
        \includegraphics[width=\linewidth]{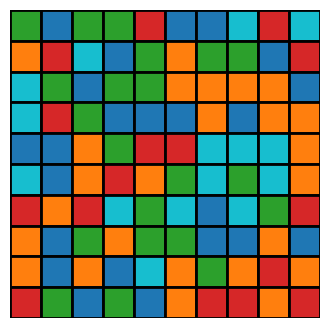}
        \caption{$10\times10\times5$}
    \end{subfigure}

    \caption{Examples of baseline images generated for this study. The grid sizes range from N=5 on the left to N=10 on the right, and the number of colors randomly assigned to each cell range from N\_colors=3 in the first row to N\_colors=5 in the third row.}
    \label{fig:baseline_image_examples}
\end{figure}

To evaluate varying degrees of spatial complexity, the grid size $N$ is incrementally scaled from $5$ to $10$, and the number of colors used was increased from $3$ to $4$ and $5$ (i.e., red, green, blue, orange, and cyan), where each cell was randomly assigned one of the possible color values (Figure~\ref{fig:baseline_image_examples}). Both the images and their corresponding exact spatial matrices are recorded to serve as the definitive baseline for all subsequent comparisons.

\subsection{Prompting LMMs}
\label{sec:prompting}

\subsubsection{Prompt 1: Image-to-Text}
\label{subsec:image-to-text}
As depicted in Figure~\ref{fig:diagram}, generated input images are provided to an LMM.
The accompanying prompt directs the LMM to act as a spatial extraction agent, tasking it to generate a machine-readable textual representation of the spatial layout that details the domain size, indexing system, and the coordinate mapping of the colored cells. The exact prompt was written as follows:

\begin{lstlisting} [caption={Prompt 1: Image-to-Text},label=lst:list1,captionpos=t,float,abovecaptionskip=-\medskipamount,breaklines=true,breakindent=0pt]
The image shows colored squares in a
grid pattern. Colors represent land use classes. Explain this image
in a text that can be used by other LMMs for spatial representation
and reasoning tasks.
\end{lstlisting}

\begin{figure}
    \centering
    \includegraphics[width=0.7\linewidth]{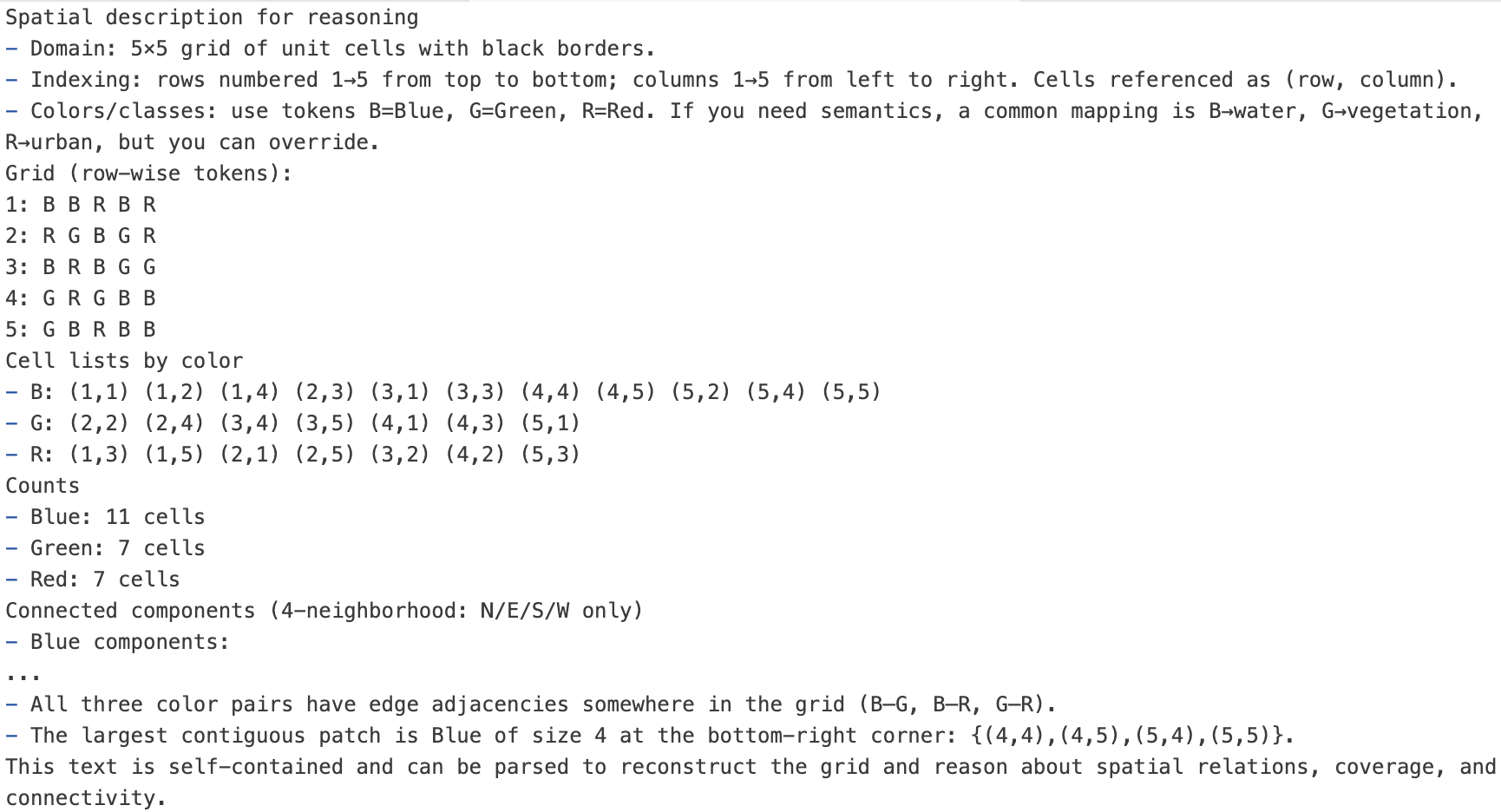}
    \caption{Textual description of the image shown in Figure~\ref{fig:553}, generated with the OpenAI's GPT5 model.}
    \label{fig:txt_response_example}
\end{figure}

As a response, the LMM would return the generated textual description of the input image. Figure~\ref{fig:txt_response_example} shows an example of the prompt response when image shown in Figure~\ref{fig:553} was used as an input.

\subsubsection{Prompt 2: Text-to-Image}
\label{subsec:text-to-image}
\begin{figure}[ht]
    \centering

    \begin{subfigure}[t]{0.155\textwidth}
        \centering
        \includegraphics[width=\linewidth]{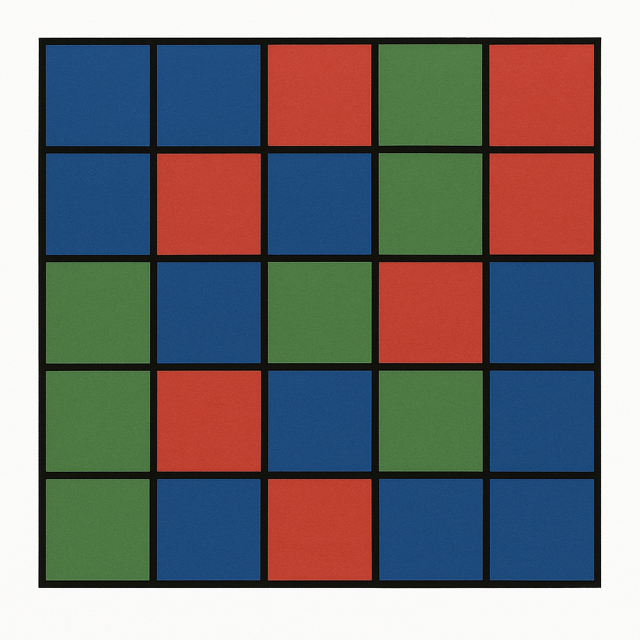}
        \caption{$5\times5\times3$}
    \end{subfigure}\hfill
    \begin{subfigure}[t]{0.155\textwidth}
        \centering
        \includegraphics[width=\linewidth]{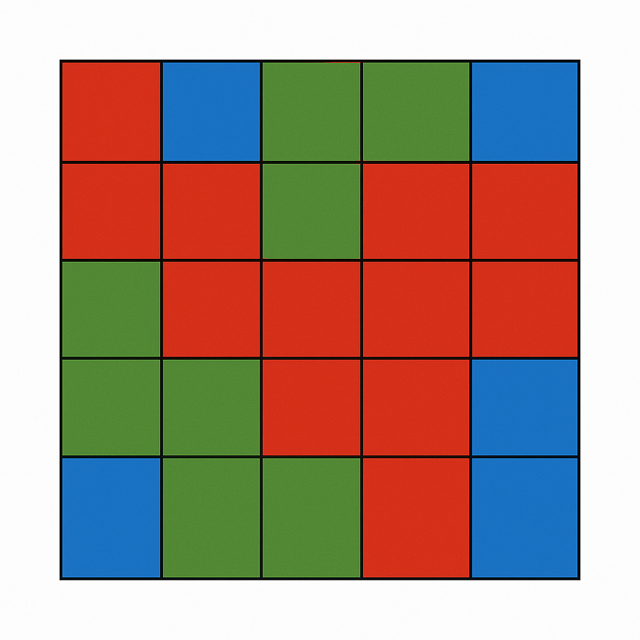}
        \caption{$6\times6\times3$}
    \end{subfigure}\hfill
    \begin{subfigure}[t]{0.155\textwidth}
        \centering
        \includegraphics[width=\linewidth]{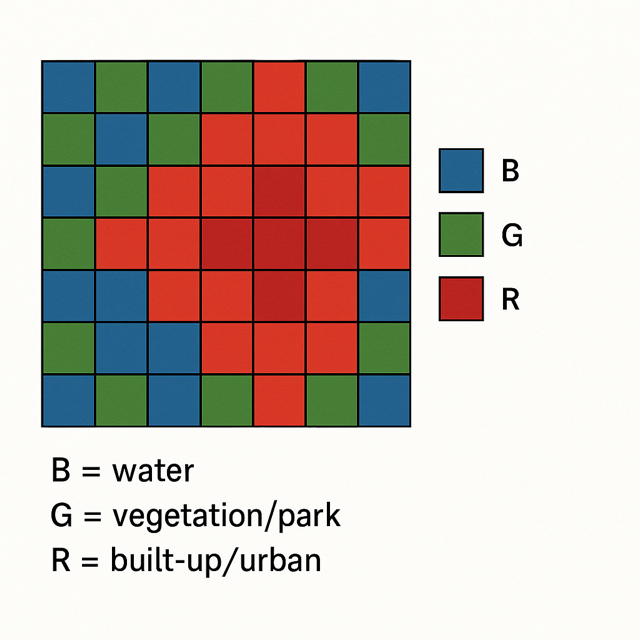}
        \caption{$7\times7\times3$}
    \end{subfigure}\hfill
    \begin{subfigure}[t]{0.155\textwidth}
        \centering
        \includegraphics[width=\linewidth]{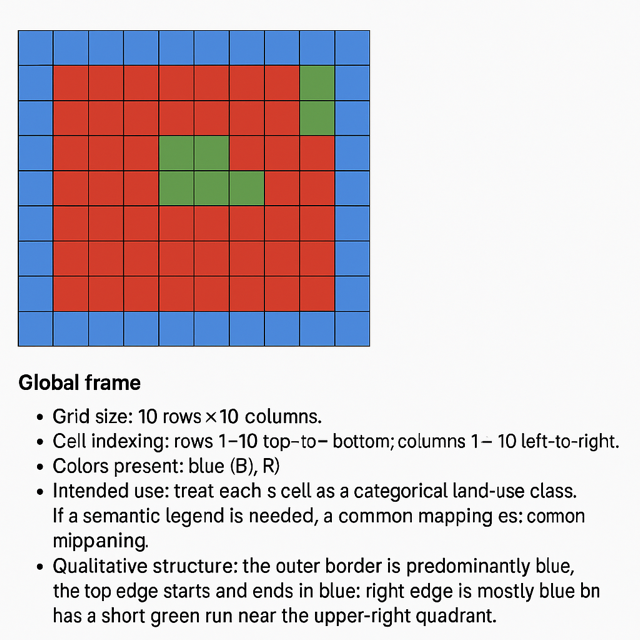}
        \caption{$8\times8\times3$}
    \end{subfigure}\hfill
    \begin{subfigure}[t]{0.155\textwidth}
        \centering
        \includegraphics[width=\linewidth]{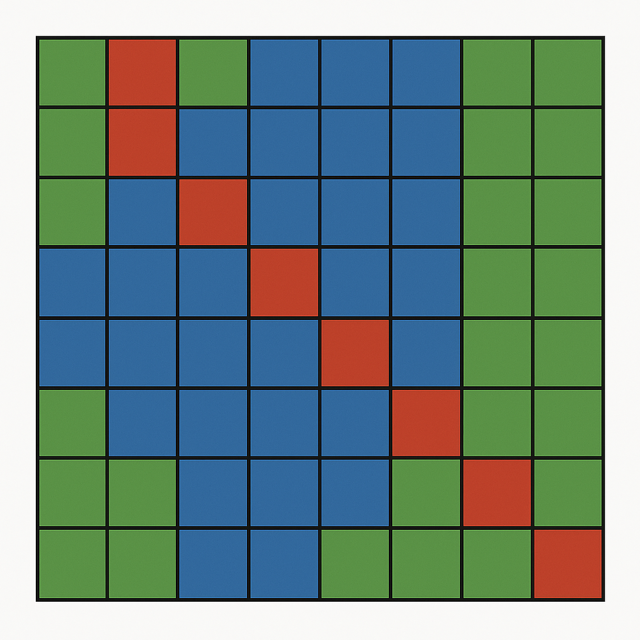}
        \caption{$9\times9\times3$}
    \end{subfigure}\hfill
    \begin{subfigure}[t]{0.155\textwidth}
        \centering
        \includegraphics[width=\linewidth]{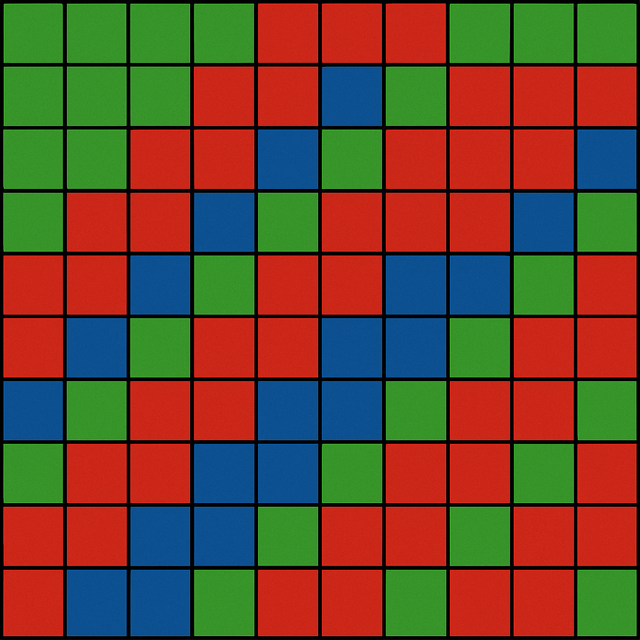}
        \caption{$10\times10\times3$}
    \end{subfigure}

    \vspace{4pt}

    \begin{subfigure}[t]{0.155\textwidth}
        \centering
        \includegraphics[width=\linewidth]{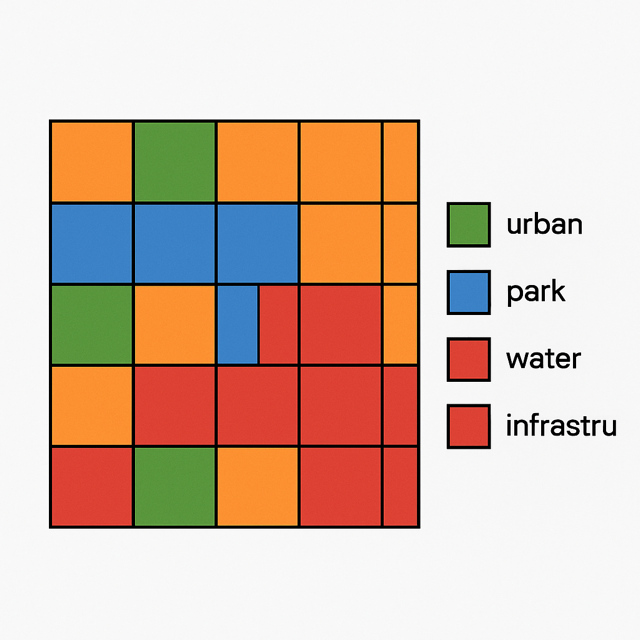}
        \caption{$5\times5\times4$}
    \end{subfigure}\hfill
    \begin{subfigure}[t]{0.155\textwidth}
        \centering
        \includegraphics[width=\linewidth]{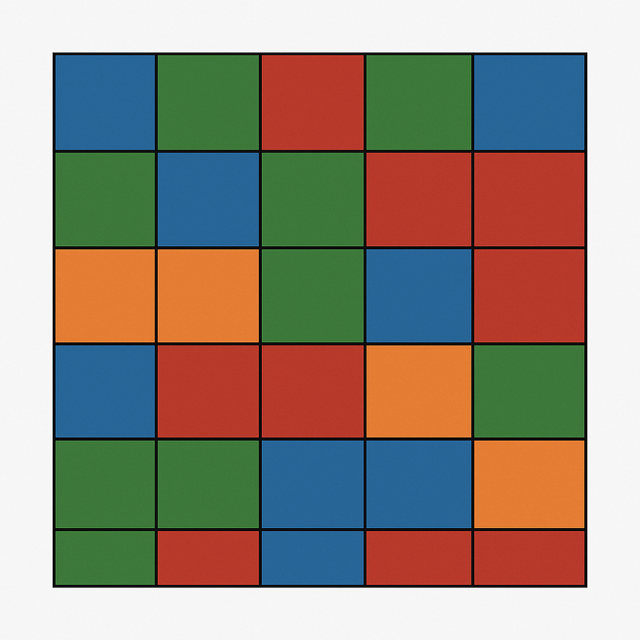}
        \caption{$6\times6\times4$}
    \end{subfigure}\hfill
    \begin{subfigure}[t]{0.155\textwidth}
        \centering
        \includegraphics[width=\linewidth]{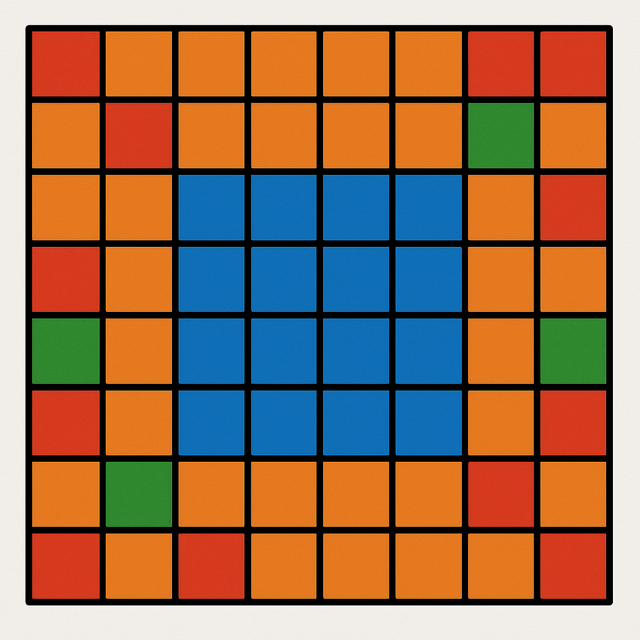}
        \caption{$7\times7\times4$}
    \end{subfigure}\hfill
    \begin{subfigure}[t]{0.155\textwidth}
        \centering
        \includegraphics[width=\linewidth]{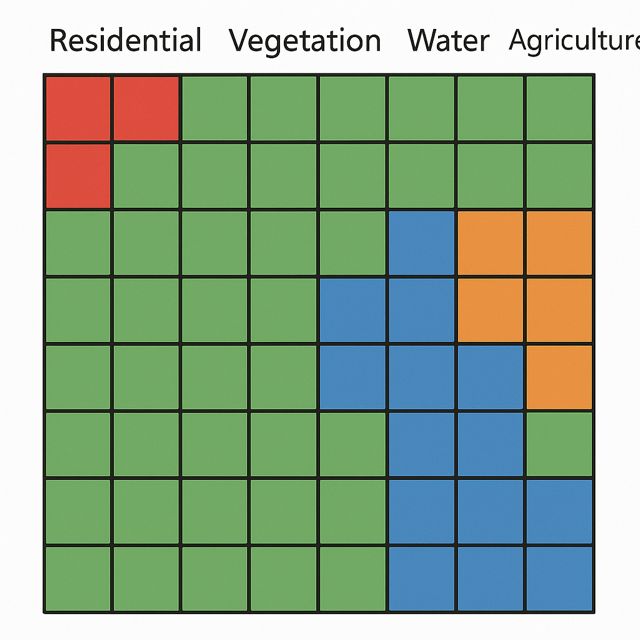}
        \caption{$8\times8\times4$}
    \end{subfigure}\hfill
    \begin{subfigure}[t]{0.155\textwidth}
        \centering
        \includegraphics[width=\linewidth]{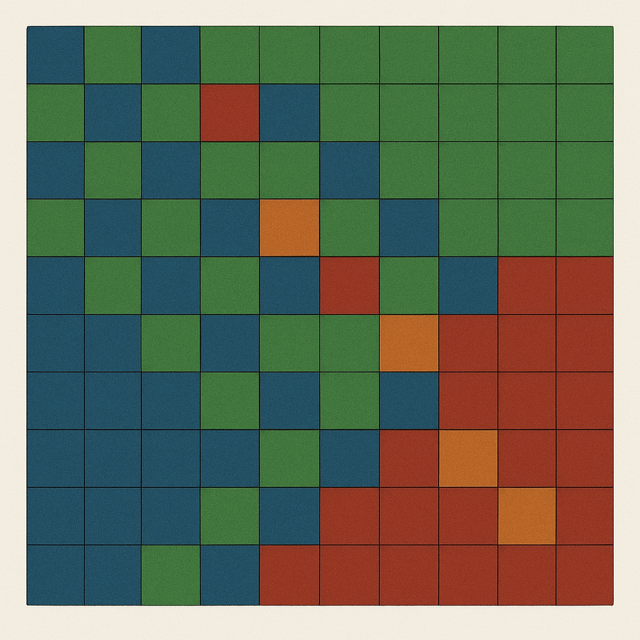}
        \caption{$9\times9\times4$}
    \end{subfigure}\hfill
    \begin{subfigure}[t]{0.155\textwidth}
        \centering
        \includegraphics[width=\linewidth]{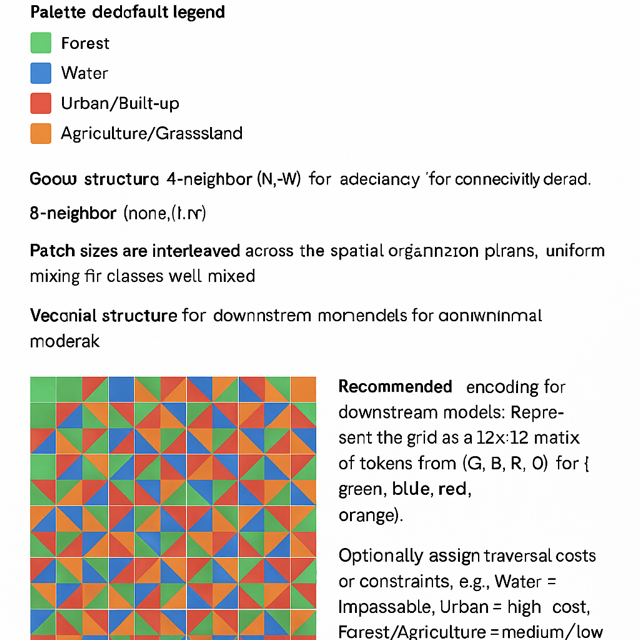}
        \caption{$10\times10\times4$}
    \end{subfigure}

    \vspace{4pt}

    \begin{subfigure}[t]{0.155\textwidth}
        \centering
        \includegraphics[width=\linewidth]{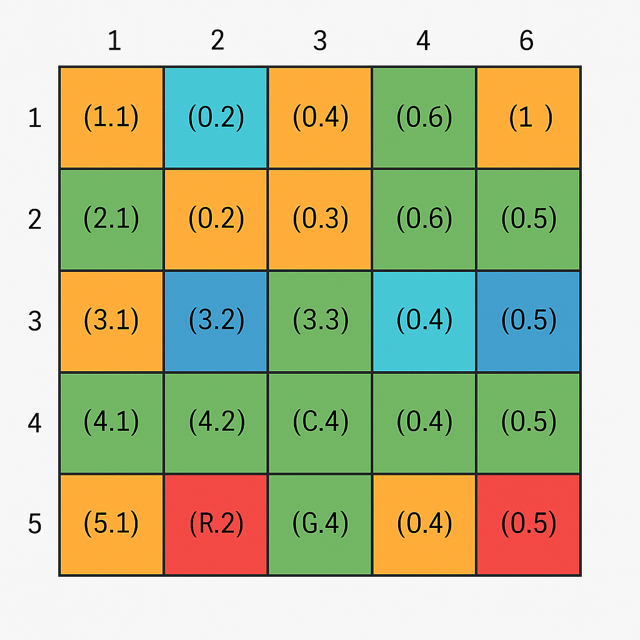}
        \caption{$5\times5\times5$}
    \end{subfigure}\hfill
    \begin{subfigure}[t]{0.155\textwidth}
        \centering
        \includegraphics[width=\linewidth]{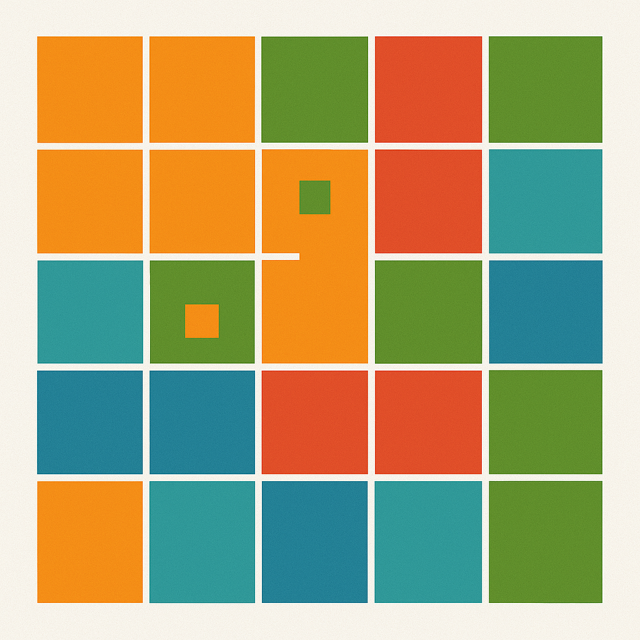}
        \caption{$6\times6\times5$}
    \end{subfigure}\hfill
    \begin{subfigure}[t]{0.155\textwidth}
        \centering
        \includegraphics[width=\linewidth]{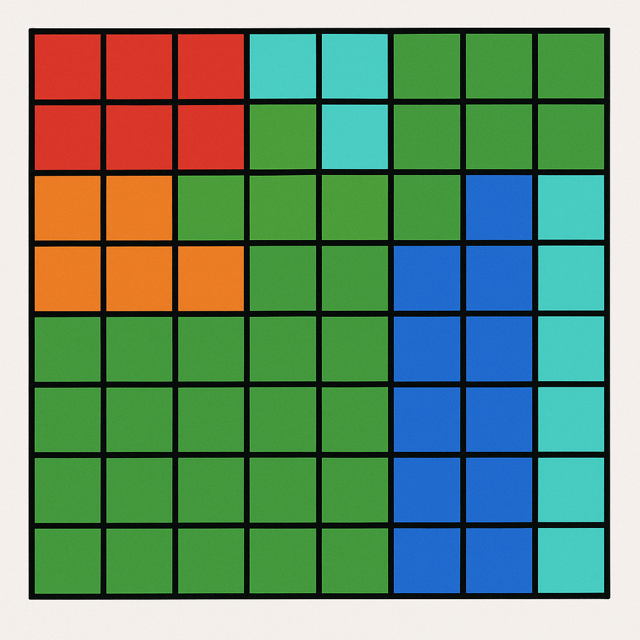}
        \caption{$7\times7\times5$}
    \end{subfigure}\hfill
    \begin{subfigure}[t]{0.155\textwidth}
        \centering
        \includegraphics[width=\linewidth]{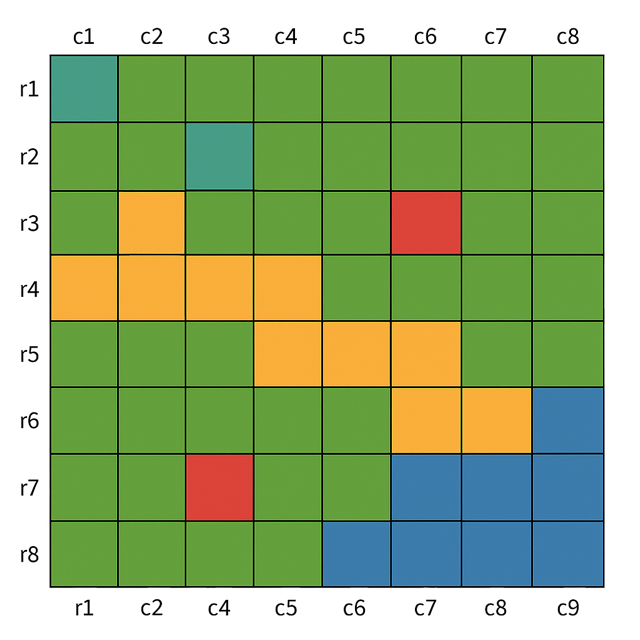}
        \caption{$8\times8\times5$}
    \end{subfigure}\hfill
    \begin{subfigure}[t]{0.155\textwidth}
        \centering
        \includegraphics[width=\linewidth]{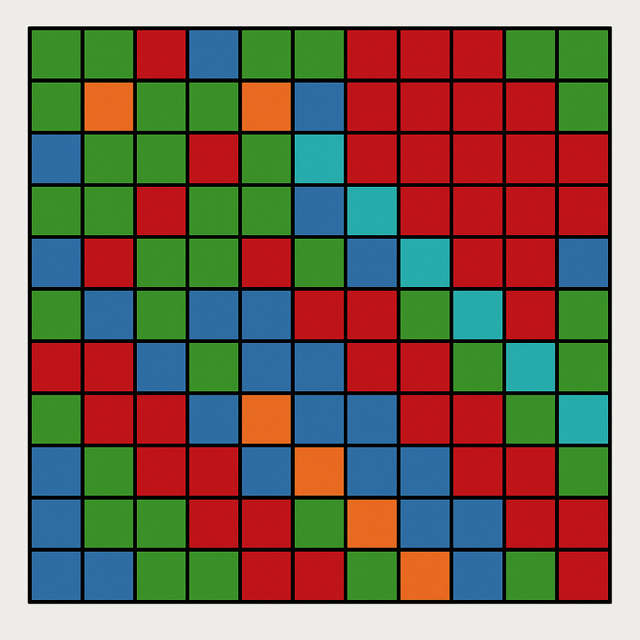}
        \caption{$9\times9\times5$}
    \end{subfigure}\hfill
    \begin{subfigure}[t]{0.155\textwidth}
        \centering
        \includegraphics[width=\linewidth]{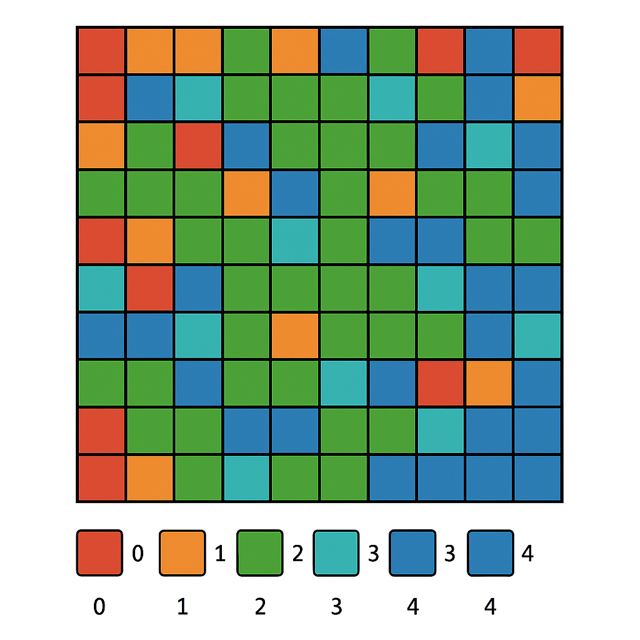}
        \caption{$10\times10\times5$}
    \end{subfigure}

    \caption{Examples of LMM generated output images based on input baseline images from Figure~\ref{fig:baseline_image_examples} after they have been through the image-to-text transfer with Prompt 1 and subsequently through the text-to-image transfer with Prompt 2.}
    \label{fig:output_image_examples}
\end{figure}

In the subsequent step, the textual description output by Prompt $1$ is used as an input for Prompt $2$ where an image-generation variant of an LMM is prompted. No additional system prompts or directions were used, as the input textual description is usually already quite detailed and the image-generation LMM understands that by default its task is to generate an image output based on the textual input. Figure~\ref{fig:output_image_examples} shows the generated output images that correspond to the input images from Figure~\ref{fig:baseline_image_examples}.

\subsection{Processing of LMM outputs}
\label{sec:processing}

To be able to compare the outputs of Prompt 1 and Prompt 2 to the input images, some preprocessing steps are required to sample and extract the relevant information and convert it into the format that allows the evaluation measures to be calculated. For this purpose, we converted all three color grid depictions - input images, generated texts, and generated images - into 2-dimensional arrays of color names. This required separate processing steps for the textual and image outputs of the LMM prompts. 

The textual description outputs of Prompt 1 (image-to-text) usually contained longer natural language descriptions of the grids visible in the input images that explained the size, format, colors used, color meanings, clusters of the same colors, and sometimes even topological relations between individual clusters. However, the majority of outputs also included some kind of matrix encoding of the input grid and this is the information that we have extracted and standardized into 2-dimensional arrays of colors where each color is represented by a single letter.

\begin{figure}
    \centering
    \includegraphics[width=0.5\linewidth]{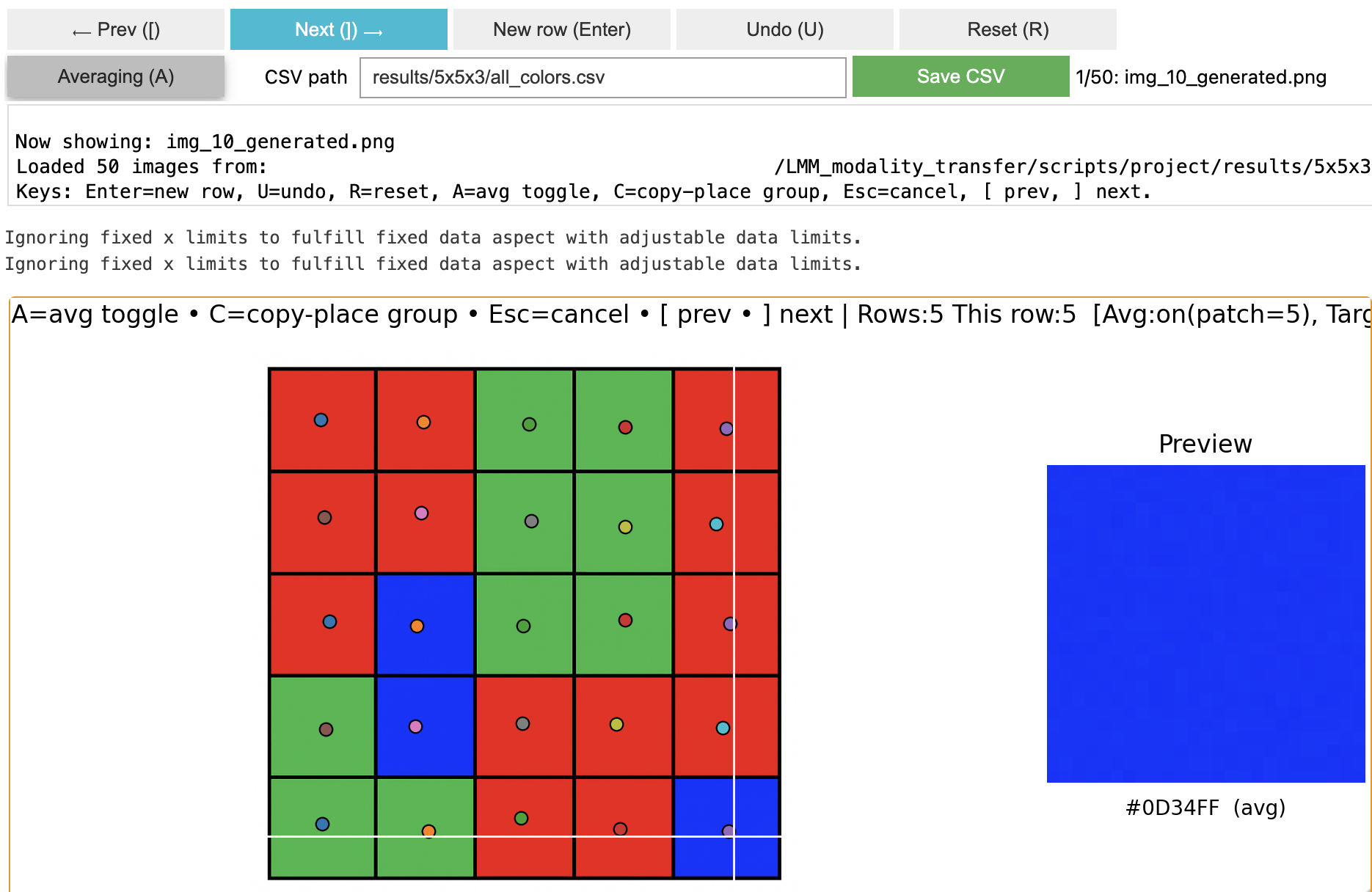}
    \caption{Graphical user interface of the output image color-sampling application. Each point represents a location where the user has sampled the cell color in order to capture the grid layout and colors of an image generated by an LMM.}
    \label{fig:color_sampling_GUI}
\end{figure}

The outputs of Prompt 2 (text-to-image) are in an image modality and require a different kind of processing to convert the information into a 2D array. Although it is possible to perform this step with yet another LMM prompt, this would be subject to the same uncertainty as Prompt 1, as it would require another image-to-text prompt and would not seem reasonable in a workflow that is intended to evaluate LMMs in the first place. Thus, we have created a small application 
that allows user to sample colors from the cells of output images using a graphical user interface (Figure~\ref{fig:color_sampling_GUI}). It should be noted that output images can also contained highly distorted grids that cannot be sampled and have to be omitted completely (see Table~\ref{tab:error_matrix}). After all output images from one directory have been sampled, the application converts the sampled points and colors into 2D arrays.


\subsection{Evaluation of LMM spatial hallucination}
\label{sec:evaluation}

Evaluating the accuracy of the modality transfer requires quantifying both spatial and attribute degradation. Generative image models rarely output exact, deterministic hex color codes, often producing slight gradients or perceptual variations. Facing this challenge, we apply a strict CIEDE2000 ($\Delta E_{00}$) thresholding function to rigorously evaluate the extracted visual matrices against the discrete ground-truth grids. Let $\mathcal{P} = \{p_1, p_2, \dots, p_k\}$ denote the ground-truth color palette. For each generated pixel color $c$, this heuristic calculates the perceptual distance $\Delta E_{00}(c, p_i)$ to every palette color. Generated colors falling within a defined perceptual threshold $\tau$ are mapped to their nearest ground-truth categorical class, while those exceeding the threshold are penalized and registered as out-of-bounds attribute hallucinations ($h$). The assigned class $l$ for a generated color $c$ is defined as:
$$
l = 
\begin{cases} 
\arg\min_{p_i \in \mathcal{P}} \Delta E_{00}(c, p_i) & \text{if } \min \Delta E_{00}(c, p_i) \leq \tau \\ 
h & \text{otherwise} 
\end{cases}
$$ 

\subsubsection{1D Levenshtein Distance}
Following their categorical normalization into discrete matrices, let $M$ represent the ground-truth matrix and $\hat{M}$ represent the generated matrix. To facilitate the evaluation of spatial determinism, these 2D spatial structures are flattened into 1D sequential strings, $S$ and $\hat{S}$, such that $S = (s_1, s_2, \dots, s_{N^2})$. Subsequently, the Levenshtein Distance \cite{levenshtein1966binary} is computed between the generated sequences and the ground truth to strictly quantify spatial loss. Let $\mathcal{L}_{S,\hat{S}}(|S|, |\hat{S}|)$ denote the distance between the two sequences, defined recursively to calculate the minimum number of mathematical edit operations required to reconstruct the original input sequence:
$$
\mathcal{L}_{S,\hat{S}}(i, j) = 
\begin{cases} 
\max(i, j) & \text{if } \min(i, j) = 0 \\ 
\min 
\begin{cases} 
\mathcal{L}_{S,\hat{S}}(i-1, j) + 1 \\ 
\mathcal{L}_{S,\hat{S}}(i, j-1) + 1 \\ 
\mathcal{L}_{S,\hat{S}}(i-1, j-1) + \mathbb{I}(s_i \neq \hat{s}_j), 
\end{cases} & \text{otherwise}
\end{cases}
$$ 
where $\mathbb{I}$ is the indicator function that equals $1$ when the characters differ and $0$ when they match. Within this spatial context, the operations directly translate to model errors: insertions ($\mathcal{L}_{S,\hat{S}}(i, j-1) + 1$) represent hallucinated blocks, deletions ($\mathcal{L}_{S,\hat{S}}(i-1, j) + 1$) represent missing blocks, and substitutions ($\mathbb{I}(s_i \neq \hat{s}_j) = 1$) represent incorrectly placed or colored blocks. If the input and output images are identical (given the allowed perceptual difference threshold for colors $\tau$), the Levenshtein Distance between them is zero, i.e., $\mathcal{L}_{S,\hat{S}}(|S|,|\hat{S}|)=0$.

\subsubsection{2D Spatial Earth Mover's Distance}
While the 1D Levenshtein sequence evaluation effectively captures topological errors of generated images and texts, this evaluation is inherently insensitive to the geometric magnitude of spatial displacements. To quantify the true 2D geographic loss and penalize severe spatial hallucinations, we formulate a discrete, class-wise Spatial Earth Mover's Distance (EMD) \cite{rubner2000earth} as an optimal transport problem.

Let $\mathcal{U}$ denote the union of all categorical labels present in both the ground-truth matrix $M$ and the generated matrix $\hat{M}$. For each label $l \in \mathcal{U}$, we extract the sets of 2D Cartesian coordinates representing the spatial locations of those specific class blocks: $C_l = \{ (x,y) \mid M_{x,y} = l \}$ and $\hat{C}_l = \{ (x,y) \mid \hat{M}_{x,y} = l \}$. Let $n_l = |C_l|$ and $\hat{n}_l = |\hat{C}_l|$ denote the cardinality of these sets.

To account for unmapped features, in particular deletions where the LMM fails to generate a required block, or insertions where the model hallucinates a new, out-of-bounds color, a maximum spatial penalty $\rho = \sqrt{H^2 + W^2}$ is defined, which represents the maximum diagonal distance of the  $H \times W$ spatial grid, where $H$ and $W$ stand for height and width of the grid.

In particular, for a given class $l$, the spatial transport cost $W(C_l, \hat{C}_l)$ computes the minimum Euclidean distance required to align the generated blocks with the ground truth, penalized by any cardinality mismatch:
$$W(C_l, \hat{C}_l) = \min_{\pi} \sum_{i=1}^{\min(n_l, \hat{n}_l)} \| c_i - \hat{c}_{\pi(i)} \|_2 + |n_l - \hat{n}_l| \rho,$$
where $\pi$ represents the optimal injective mapping from the smaller coordinate set to the larger coordinate set, solved via the Hungarian algorithm (linear sum assignment), and $\| \cdot \|_2$ denotes the standard $L_2$ Euclidean distance. Finally, the aggregated Spatial EMD is divided by the total number of valid ground-truth blocks, $N_{gt} = \sum_{l \in \mathcal{U}} n_l$, which ensures that the final metric represents the average physical pixel shift per correct block:
$$\text{EMD} = \frac{1}{N_{gt}} \sum_{l \in \mathcal{U}} W(C_l, \hat{C}_l).$$


\subsubsection{Interpretation of Metrics}
To systematically evaluate the modality transfer pipeline and diagnose the exact nature of spatial degradation, we analyze the interplay between the 1D sequence error (i.e., 1D Levenshtein) and the 2D geometric error (i.e., 2D spatial EMD). While the Levenshtein distance effectively measures whether the generative model understood the topological shape and categorical sequence of the prompt, it is highly sensitive to minor structural shifts and heavily penalizes simple translations. Conversely, the spatial EMD quantifies the absolute geographic magnitude of these displacements, indicating whether the model adheres to strict coordinate geometry.

By evaluating these two metrics, as summarized in Table \ref{tab:error_matrix}, we establish a diagnostic framework to classify the spatial reasoning failure of LMMs. This decoupled approach allows us to definitively distinguish between minor spatial coordinate drifts, where the model maintains relative topology but fails at absolute positioning and catastrophic generative breakdowns, where all spatial context is completely hallucinated.

\begin{table}[htbp]
    \centering
    \caption{Spatial Error Diagnostic Matrix based on Levenshtein Distance and Spatial EMD.}
    \label{tab:error_matrix}
    \begin{tabular}{@{}llp{9cm}@{}}
        \toprule
        \textbf{Levenshtein} & \textbf{Spatial EMD} & \textbf{Diagnostic Category \& Interpretation} \\
        \midrule
        Low & Low & \textbf{Minor Attribute Error:} The model successfully preserves sequence and geometry. Represents flawless transfer (if both are exactly 0) or perfect spatial layouts with minor categorical substitutions (e.g., incorrect colors). \\
        \addlinespace
        High & Low & \textbf{Slight Misalignment (Topology Failure):} The model preserves general shape and clustering but lacks absolute coordinate precision. Minor translations (e.g., a 1-cell shift) heavily penalize the 1D sequence while keeping 2D geometric loss low. \\
        \addlinespace
        High & High & \textbf{Severe Hallucination (Catastrophic Failure):} A complete loss of geometric and sequential context. Generated blocks are scattered randomly, omitted entirely, or rendered with out-of-bounds, hallucinated colors. \\
        \addlinespace
        Low & High & \textbf{Geometric Outlier (Shape Distortion):} A mathematically anomaly where the 1D sequence remains mostly intact, but physical boundary violations (e.g., hallucinating a "jagged" grid) incur a massive maximum-diagonal geometric penalty. \\
        \bottomrule
    \end{tabular}
\end{table}

\section{Results}
\label{sec:result}

We performed an experiment where recent generation OpenAI LMMs were evaluated with the modality transfer task proposed in Section~\ref{sec:approach}. We first generated input images with the grid sizes ($N\times N$) ranging from $N=5$ to $N=10$, and the number of colors ranging from $N_{colors}=3$ to $N_{colors}=5$. For each parameter combination (e.g., $5\times 5\times 3$), $50$ distinct baseline images were created, resulting in a total of $900$ input images. For Prompt 1 (image-to-text), we utilize OpenAI's \texttt{gpt-5} model\footnote{\url{https://developers.openai.com/api/docs/models/gpt-5}} to generate textual descriptions of input images with $temperature=1$ and default token size. For Prompt 2 (text-to-image), we use OpenAI's \texttt{gpt-image-1} model\footnote{\url{https://developers.openai.com/api/docs/models/gpt-image-1}} to generate new images based on the outputs of Prompt 1. We set the parameters for Prompt 2 as $image\_size=1024\times 1024$ (pixels) and $quality=high$. It is important to note that OpenAI's API is a paid service and image generation prompts may accumulate particularly high costs. We executed prompts for $N_{colors}=3$ cases by sending API calls directly from our Python scripts. For all other cases, we executed prompts in batches directly on the OpenAI's API dashboard in order to lower the costs and save resources. Total cost of the entire experiment --- $900$ \texttt{gpt-5} prompts, $900$ \texttt{gpt-image-1} prompts, as well as a number of test prompts --- was approximately $200$ USD.

\begin{figure}[htbp]
    \centering

    \begin{subfigure}[b]{1\textwidth} 
        \centering
        \includegraphics[width=\textwidth]{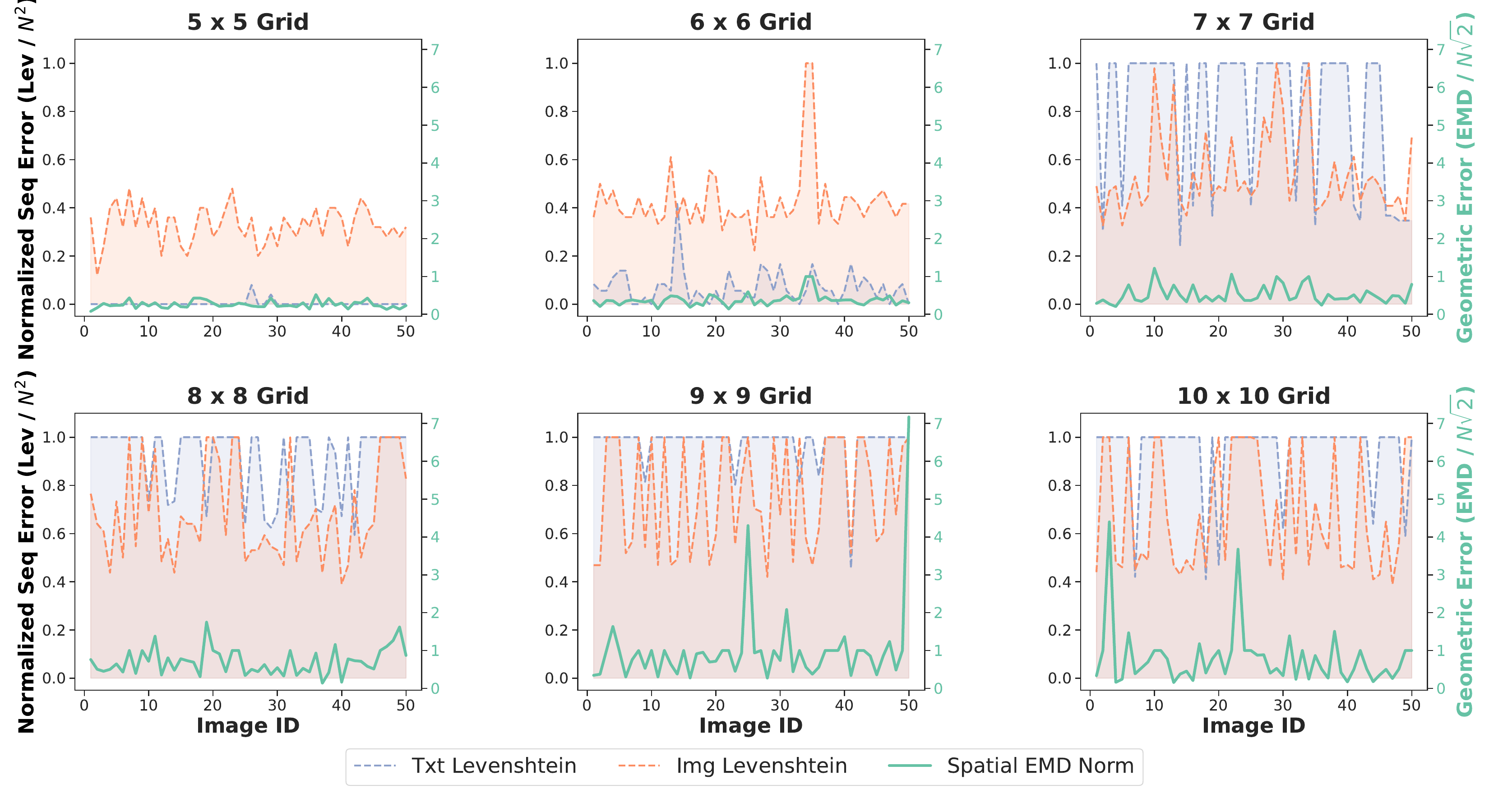}
        \caption{Evaluation metrics for spatial grids utilizing a \textit{\textbf{3-color palette}} (i.e., red, green, blue). At this lower number of grid and color complexity, the models maintain relatively stable topological and geometric coherence at smaller grid sizes ($N \le 7$). While slight misalignments occur, catastrophic geometric failures (high EMD) remain infrequent until the grid size reaches $N \ge 9$.}
        \label{fig:spatial_metrics_visualization_3_colors}
    \end{subfigure}

    \begin{subfigure}[b]{1\textwidth}
        \centering
        \includegraphics[width=\textwidth]{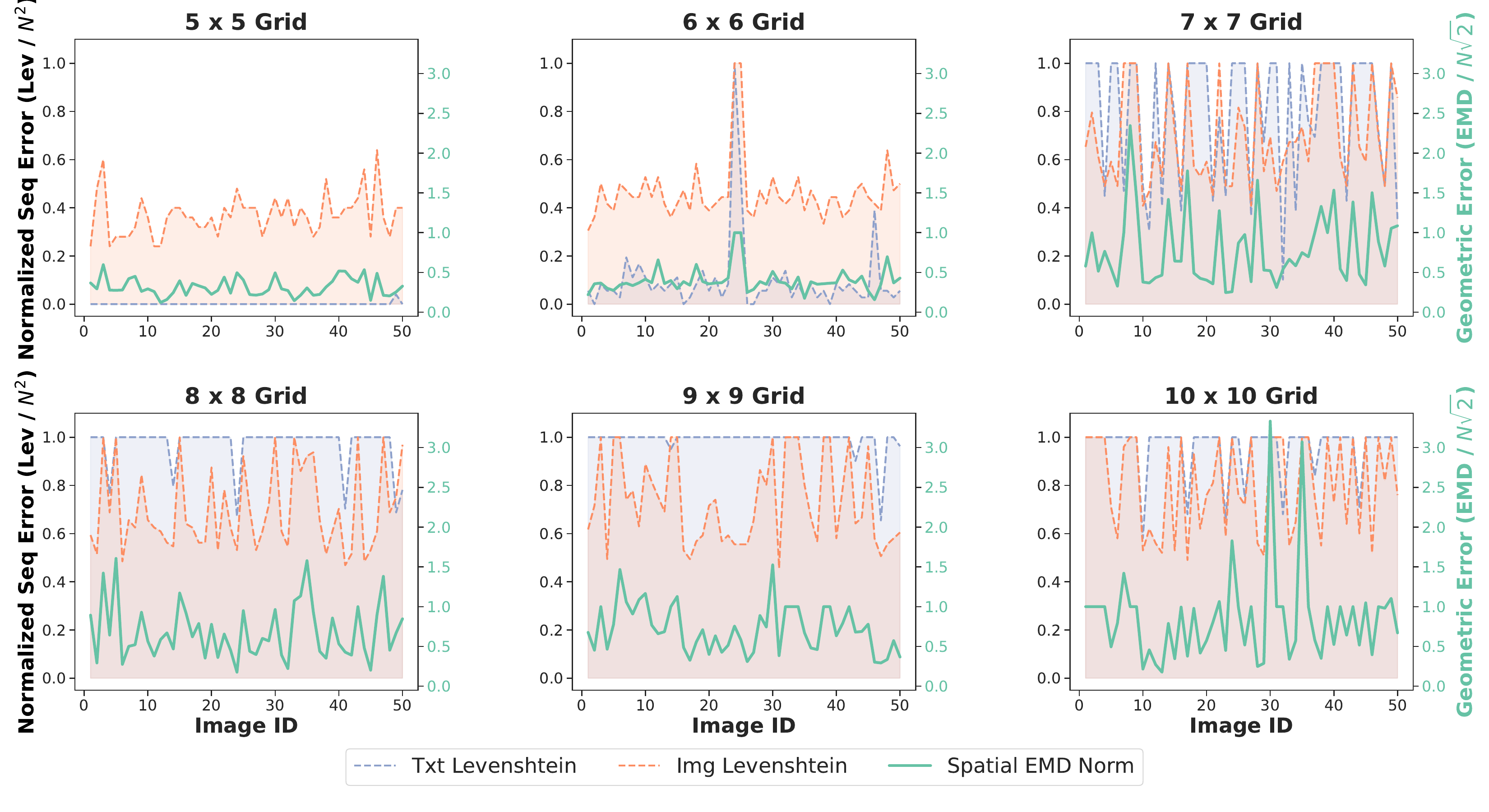}
        \caption{Evaluation metrics for spatial grids utilizing a \textit{\textbf{4-color palette}} (i.e., red, green, blue, orange). The introduction of a fourth categorical color significantly destabilizes the modality transfer. Notably, at $N \ge 8$, the models exhibit simultaneous, severe spikes in both sequence error (Img Levenshtein) and geometric displacement (Spatial EMD), indicating complete losses of spatial context within individual samples.}
        \label{fig:spatial_metrics_visualization_4_colors}
    \end{subfigure}
\end{figure}

\begin{figure}[htbp]
    \ContinuedFloat %
    \centering
    \begin{subfigure}[b]{1\textwidth}
        \centering
        \includegraphics[width=\textwidth]{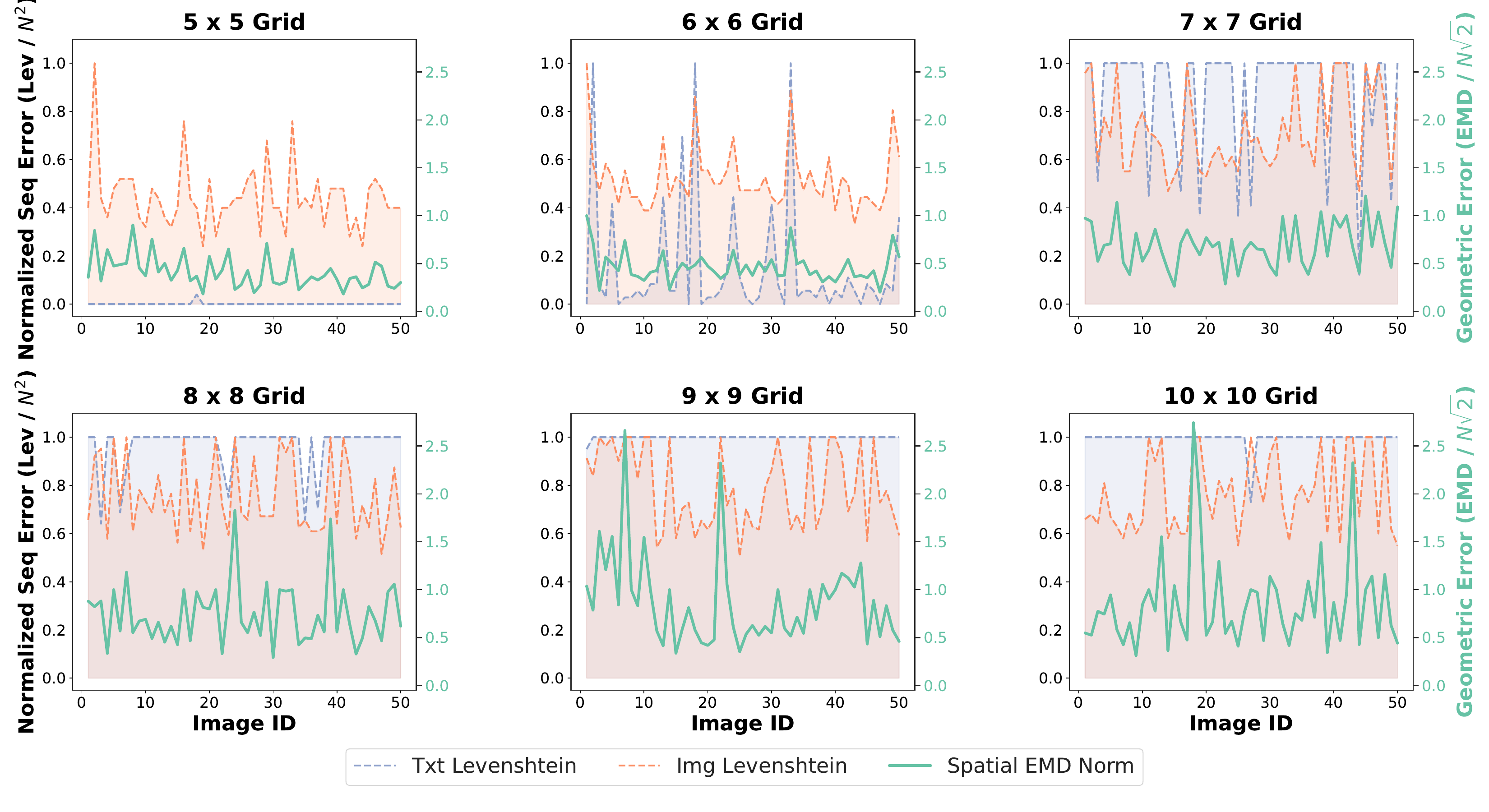}
        \caption{Evaluation metrics for spatial grids utilizing a \textit{\textbf{5-color palette}} (i.e., red, green, blue, orange, cyan). Maximum visual complexity results in immediate and significant modality transfer breakdowns across nearly all grid sizes. The baseline vision-to-text parsing (Txt Levenshtein) exhibits frequent failures early on, subsequently corrupting the text-to-image generation phase and driving continuous catastrophic geometric errors.}
        \label{fig:spatial_metrics_visualization_5_colors}
    \end{subfigure}
    
    \caption[]{Detailed evaluation of modality transfer performance across varying grid sizes ($N=5$ to $N=10$) and color palette complexities (3, 4, and 5 colors). The primary y-axis (left) displays the normalized 1D sequence error (Levenshtein distance divided by $N^2$), contrasting the intermediate text description error (blue dotted line) against the final generated image error (red shaded line). The secondary y-axis (right) displays the 2D geometric error via normalized Spatial EMD (green solid line). Across all configurations, scaling the grid dimensions and increasing the number of categorical colors correlates with an exponential rise in spatial hallucinations and loss of geometric determinism.}

    \label{fig:spatial_metrics_visualization}
\end{figure}

\begin{figure}[htbp]
    \centering
    
    \begin{subfigure}[t]{0.5\textwidth} 
        \centering
        \includegraphics[width=\textwidth]{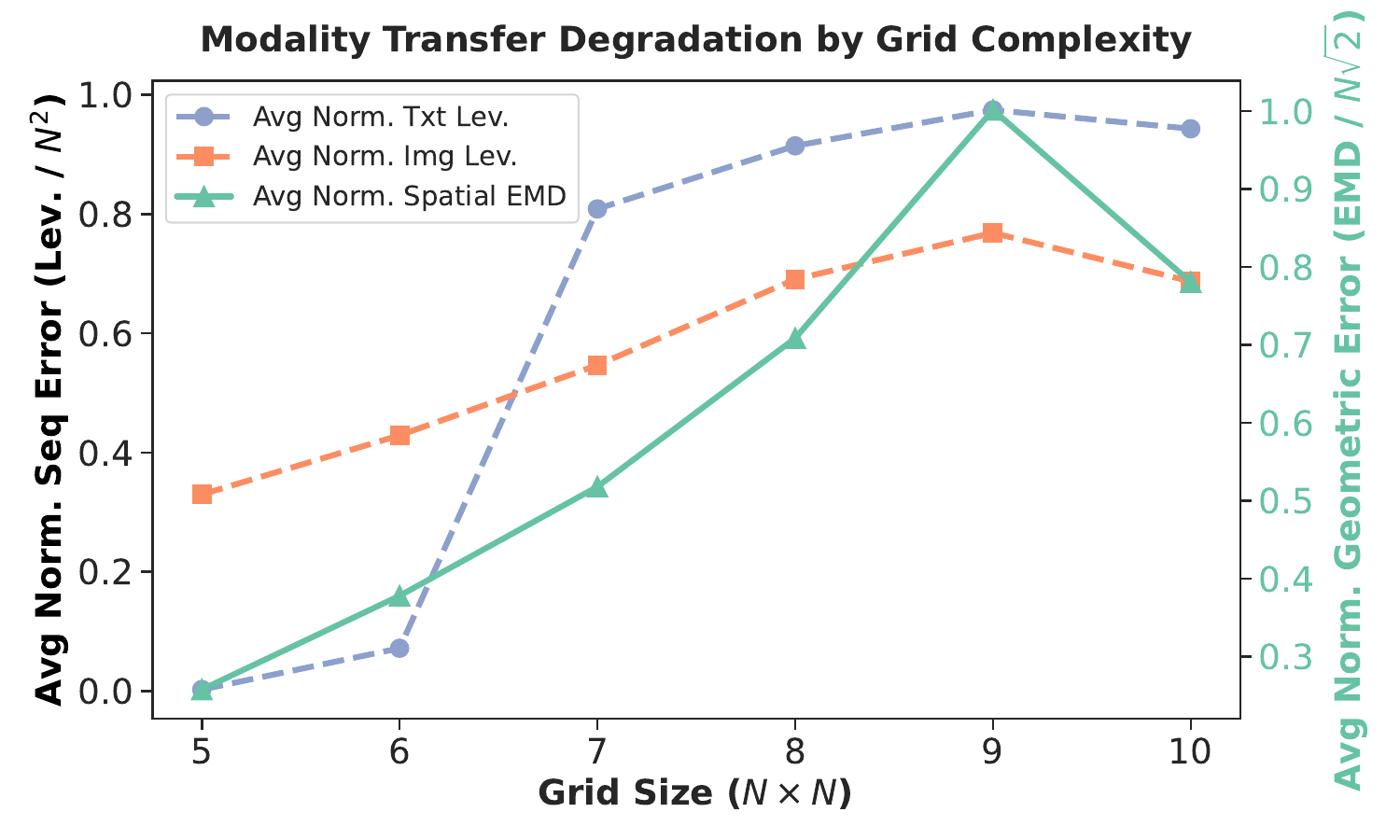}
        \caption{\textit{\textbf{3-color palette}}}
        \label{fig:average_error_scaling_3_colors}
    \end{subfigure}\hfill
    \begin{subfigure}[t]{0.5\textwidth}
        \centering
        \includegraphics[width=\textwidth]{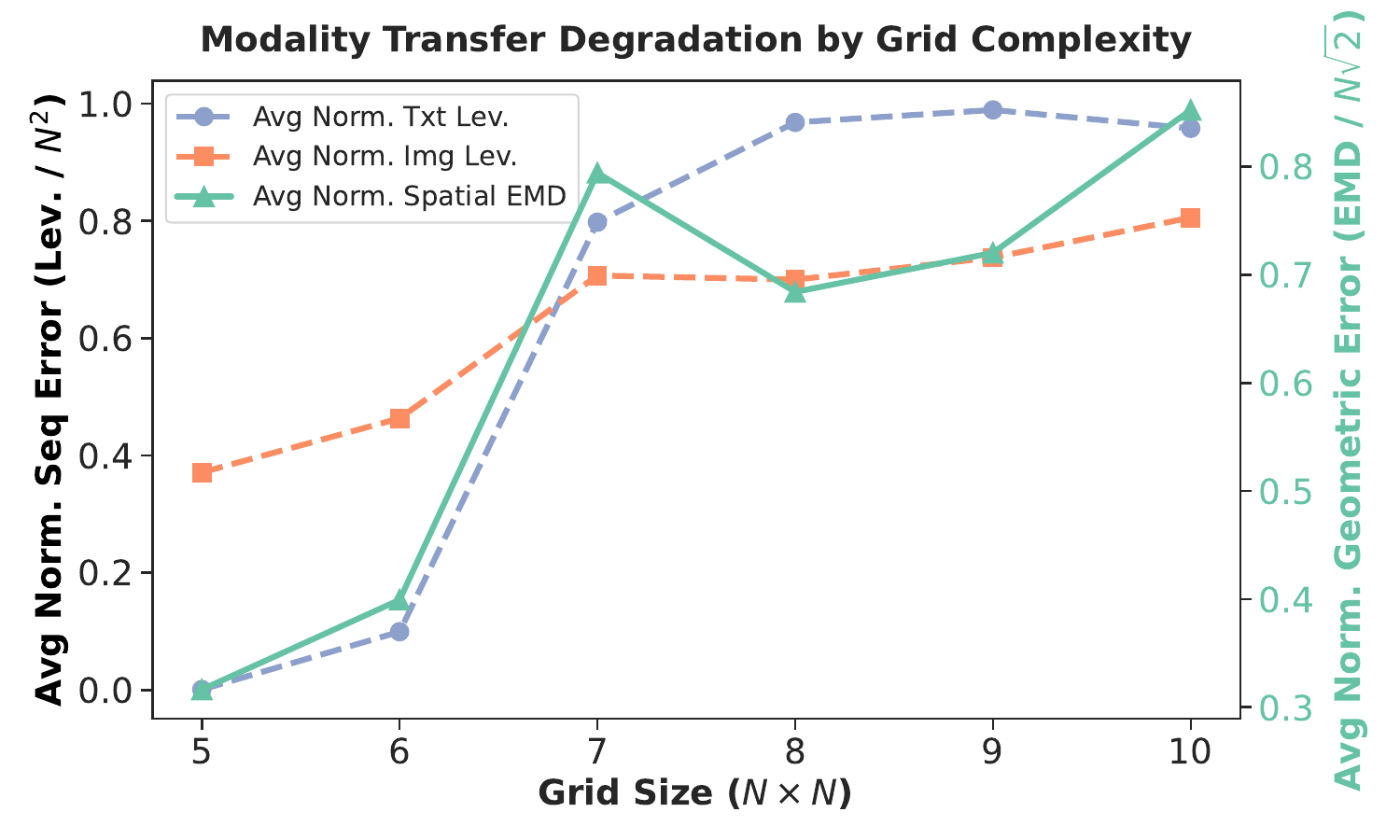}
        \caption{\textit{\textbf{4-color palette}}}
        \label{fig:average_error_scaling_4_colors}
    \end{subfigure}
        
    \begin{subfigure}[t]{0.5\textwidth}
        \centering
        \includegraphics[width=\textwidth]{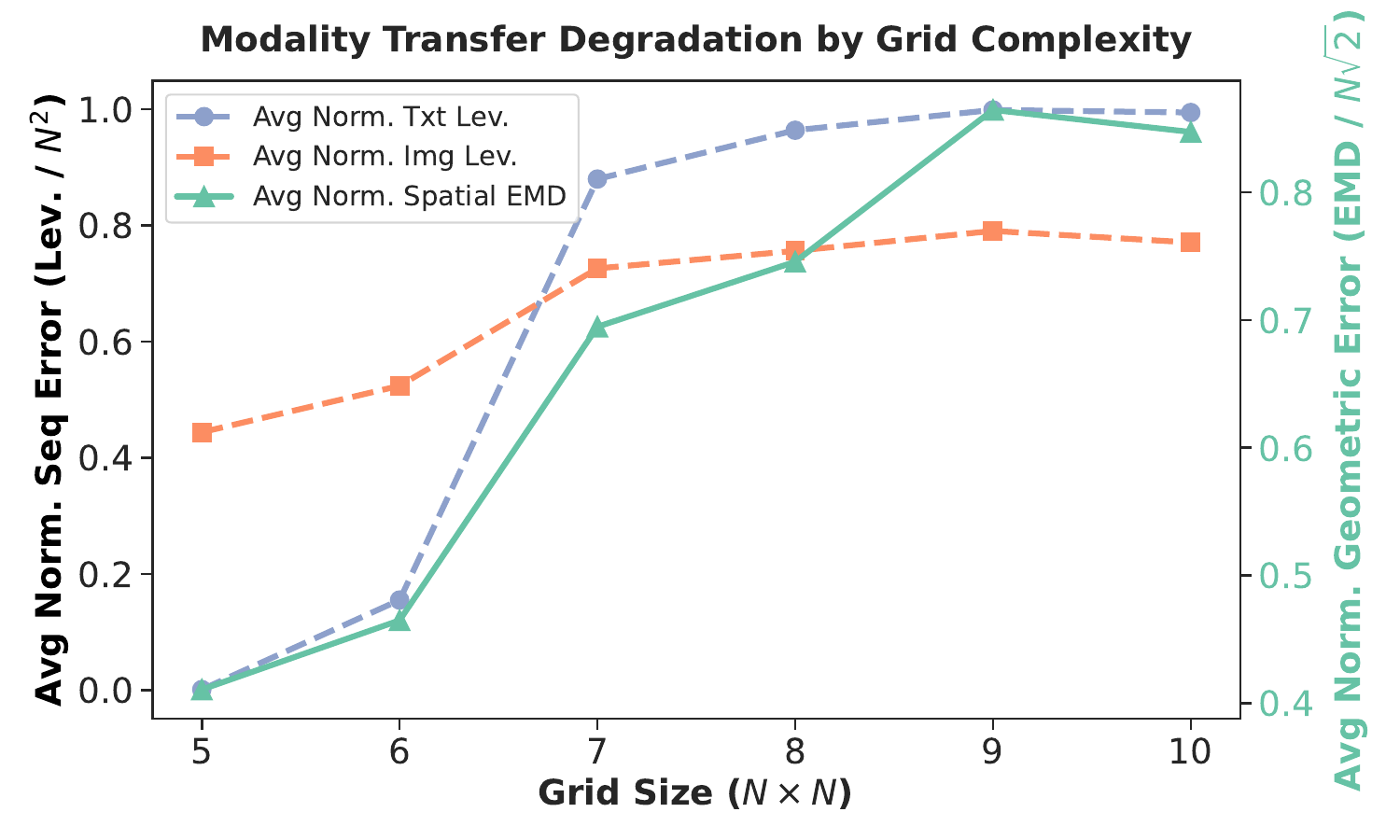}
        \caption{\textit{\textbf{5-color palette}}}
        \label{fig:average_error_scaling_5_colors}
    \end{subfigure}
    
    \caption{Scaling behavior of average spatial reasoning errors across varying grid dimensions ($N=5$ to $N=10$). The left y-axis displays the normalized 1D sequence errors (Levenshtein distance divided by $N^2$), tracking topological failures in the text description phase (blue dashed line) and image generation phase (red dashed line). The right y-axis tracks 2D geometric error via normalized Spatial EMD (green solid line). Individual figures show different trends in these measures when performing modality transfer for cases with 3 \textbf{(a)}, 4 \textbf{(b)}, and 5 colors \textbf{(c)}.}
    \label{fig:average_error_scaling}
\end{figure}

Figure~\ref{fig:spatial_metrics_visualization} shows the evaluation of each individual input image (Image ID axis) with the Normalized Sequential Error (Levenshtein Distance divided by grid size $N^2$) and the Normalized Geometric Error (Spatial EMD divided by grid diagonal $N\sqrt{2}$) measures. The tested models demonstrate a high degree of variability dependent on both the scale of the grid ($N$) and the color complexity. 
In the simplest configuration, testing with 3 color palette (Figure~\ref{fig:spatial_metrics_visualization_3_colors}), the vision-to-text sequence error remains relatively low for $N \le 6$. However, it exhibits a sharp capability drop at $N=7$, driving downstream errors that peak at $N=9$. Expanding the palette to 4 colors (Figure~\ref{fig:average_error_scaling_4_colors}) complicates the representational mapping and accelerates this breakdown; baseline vision parsing degrades much earlier, compounding sequence penalties and geometric misalignments. At maximum complexity with 5 colors (Figure~\ref{fig:average_error_scaling_5_colors}), the tested LMMs experience immediate catastrophic failures, where converging high sequence penalties and very high Spatial EMD values indicate a total loss of topological and geometric structure. Across all configurations, scaling spatial density consistently triggers an abrupt loss of geometric determinism.

Figure~\ref{fig:average_error_scaling} shows the aggregated average measures for each grid size and color configuration to highlight the effect of increasing spatial complexity on the performance of the LMM transfer task. Overall, the average scaling trends also identify a capability drop after a certain point.
For the 3-color palette (Figure~ \ref{fig:average_error_scaling_3_colors}), this threshold occurs abruptly at $N=7$, where the textual encodings begin to fail, cascading into significantly higher image generation errors. This phenomenon is exacerbated in the 4-color palette (Figure~\ref{fig:average_error_scaling_4_colors}), where the modality transfer success destabilizes earlier and more severely, resulting in a steady, proportional increase in both sequence and geometric errors as the grid size scales. At the maximum complexity of 5-color palette (Figure~\ref{fig:average_error_scaling_5_colors}), the models exhibit systemic failure. The initial image-to-text extraction phase (Txt Levenshtein) incurs massive errors at grid size at $N=7$. As the foundational spatial description is intrinsically corrupted, the subsequent text-to-image generation is fed hallucinated or highly disjointed spatial coordinates. This results in maximum-penalty geometric displacements (Spatial EMD) peaking consistently across $N \geq 7$, demonstrating that the models entirely lost the capacity to deterministically map or reconstruct the original spatial context.

\begin{figure}[htbp]
    \centering
    
    \begin{subfigure}[t]{0.5\textwidth} 
        \centering
        \includegraphics[width=\textwidth]{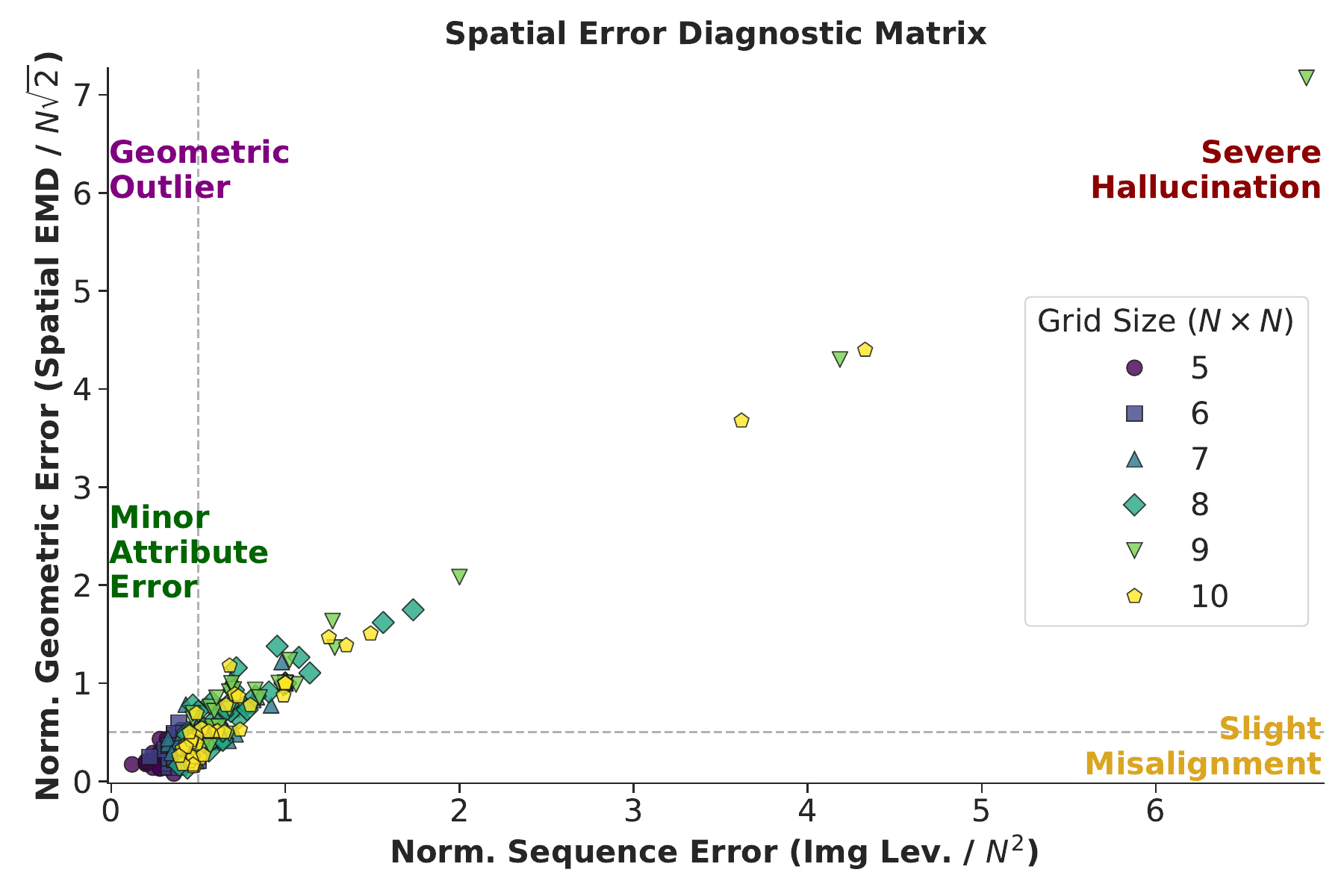}
        \caption{\textit{\textbf{3-color palette}}}
        \label{fig:diagnostic_scatter_matrix_3_colors}
    \end{subfigure}\hfill
    \begin{subfigure}[t]{0.5\textwidth}
        \centering
        \includegraphics[width=\textwidth]{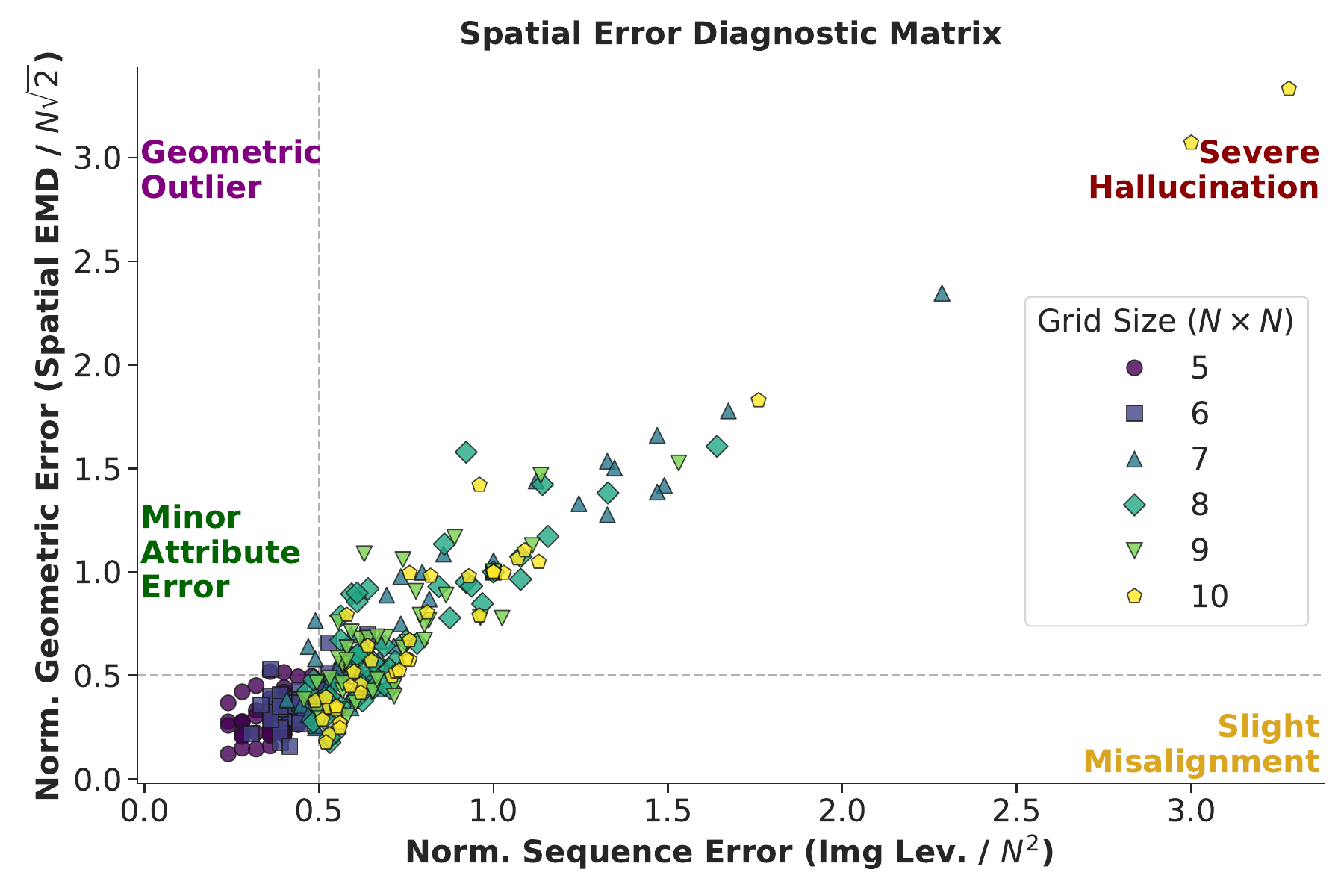}
        \caption{\textit{\textbf{4-color palette}}}
        \label{fig:diagnostic_scatter_matrix_4_colors}
    \end{subfigure}
        
    \begin{subfigure}[t]{0.5\textwidth}
        \centering
        \includegraphics[width=\textwidth]{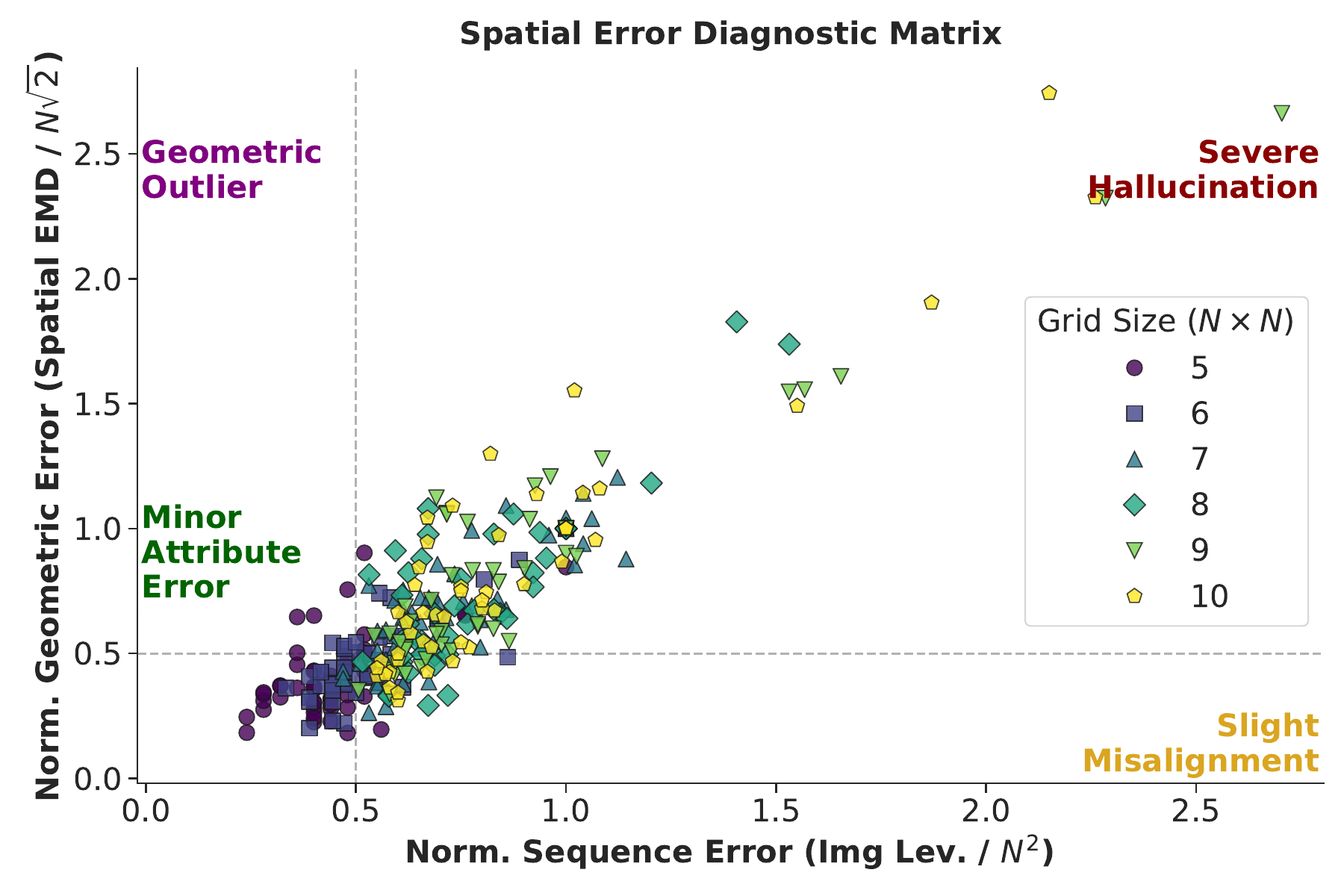}
        \caption{\textit{\textbf{5-color palette}}}
        \label{fig:diagnostic_scatter_matrix_5_colors}
    \end{subfigure}
    
    \caption{Diagnostic scatter matrices mapping 1D sequence error (Normalized Levenshtein, x-axis) against 2D geometric error (Normalized Spatial EMD, y-axis). The spatial distribution patterns directly correspond to the diagnostic categories defined in Table \ref{tab:error_matrix}. Individual figures show diagnostic matrices for different cases with 3 \textbf{(a)}, 4 \textbf{(b)}, and 5 colors \textbf{(c)}.}
    \label{fig:diagnostic_scatter_matrix}
\end{figure}

Figure~\ref{fig:diagnostic_scatter_matrix} places the individual cases into four main evaluation categories  introduced in Table~\ref{tab:error_matrix}, namely the Minor Attribute Error, Slight Misalignment (Topology Failure), Severe Hallucination (Catastrophic Failure), and Geometric Outlier (Shape Distortion). Each instance of an input image - generated text - and generated image (e.g., $5\times5\times3$ - img\_1) is plotted as a single point using the Normalizes Sequence Error and Normalized Geometric Error as the axes.
For the 3-color palette, the majority of points cluster in the bottom-left quadrant (Minor Attribute Error), indicating successful modality transfer at smaller grid scales. The 4-color palette (Figure~\ref{fig:diagnostic_scatter_matrix_4_colors}) induces broader dispersion, with a noticeable migration of $N \ge 8$ data points into the right hemisphere (Slight Misalignment), reflecting topological failures where sequences are corrupted despite partially intact geometries. At 5 colors (Figure~\ref{fig:diagnostic_scatter_matrix_5_colors}), the distribution is severely skewed toward the top-right quadrant (Severe Hallucination). This dense clustering highlights a catastrophic failure in modality transfer, where the model simultaneously loses both sequential topology and geometric precision. Overall, the progression visually demonstrates the LMM's failure trajectory as both grid scale and categorical color density overwhelm its spatial reasoning capabilities.

\section{Discussion}
\label{sec:disscussion}

While the overall results of the study are presented in Section~\ref{sec:result}, some interesting insights that were observed during the experiments are not visible from the figures and information presented there. In the following, we discuss our observations and thoughts regarding the proposed approach and the LMM modality transfer. Namely, we discuss their issues in performing image-to-text prompts and the anomalies we observed in the results of text-to-image prompts. We also provide further interpretation of the diagnostic analysis of modality transfer and lay out the limitations of our approach and case study.

\subsection{Issues in Prompt 1 (Image-to-Text) outputs}

\begin{figure}
    \centering
    \includegraphics[width=0.5\linewidth]{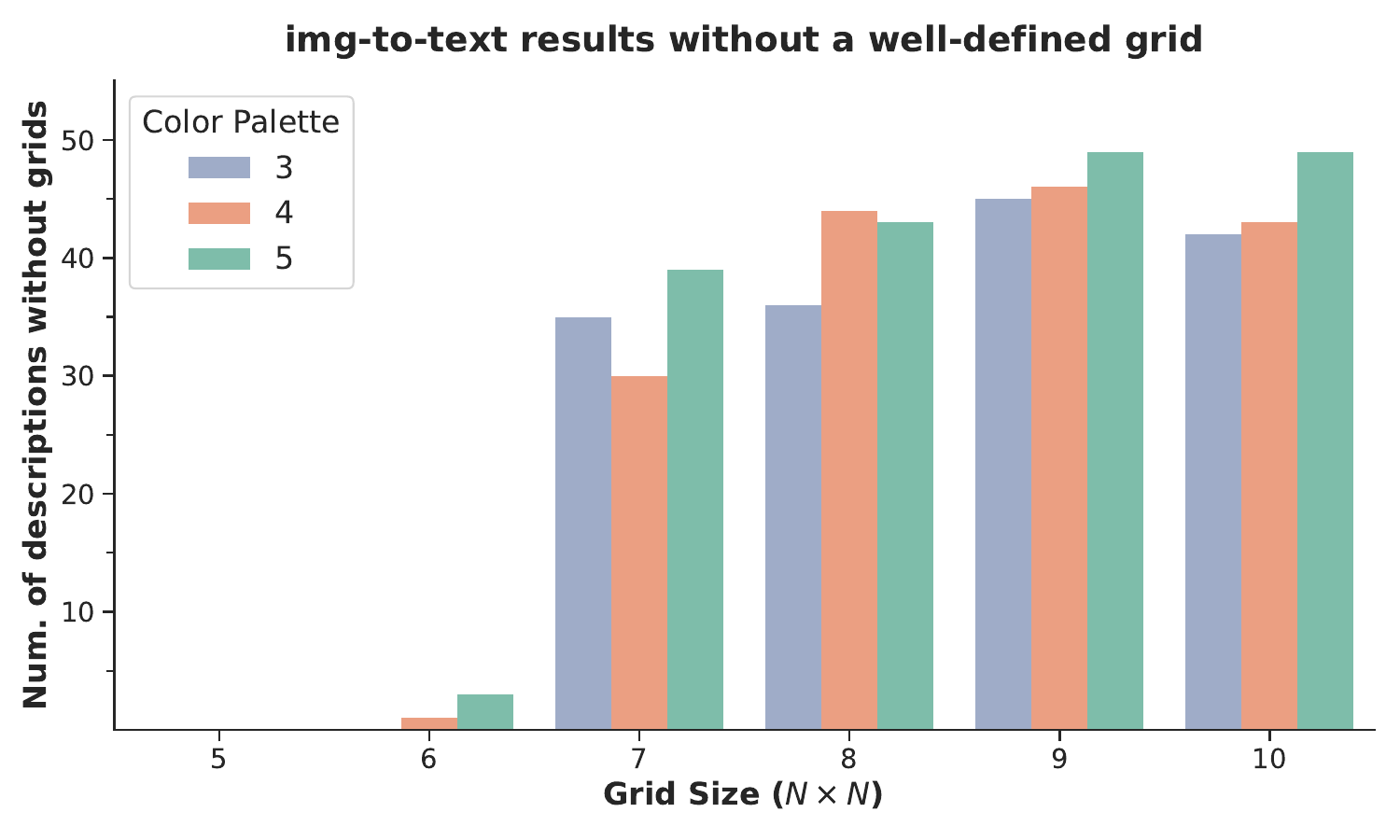}
    \caption{Bars show the number of invalid Prompt 1 (img-to-text) responses where the explicitly defined grid of colors is missing from the generated textual description. Each bar shows this value for one $N\times N\times N_{colors}$ configuration where a total of 50 prompts were executed.}
    \label{fig:txt_outputs_breakdown}
\end{figure}

We intentionally defined the instructions to LMM in Prompt 1 loosely, not to affect the spatial comprehension of the LMM with our prompt engineering. That said, the outputs for our initial input images of $5\times 5$ grids with $3$ colors seemed quite uniform as the model would briefly describe the overall task in natural language and then provide some kind of matrix representation of the grid in the input image - often with single letters representing different colors (see the example response in Figure~\ref{fig:txt_response_example}). However, when the complexity of the input grid would increase through higher cell or color count, we noticed that textual responses would sometimes not include the full grid representation at all and sometimes only include the first few rows with the note such as \emph{``If you need a lossless, per-cell matrix for this exact image, ask for per-cell encoding and I will return a row-major 2D array''}. We noticed that this effect started occuring more often from the $N=7$ and upwards -- Figure~\ref{fig:txt_outputs_breakdown} show the number of Prompt 1 responses per configuration (out of 50) where the explicit grid is missing.

If we compare this with the evaluation of the higher dimension grid cases ($N=8,9,10$) in Figure~\ref{fig:spatial_metrics_visualization}, we can notice that the images generated using such textual descriptions (i.e., cases where Txt Levenshtein $=1$ and Img Levevenshtein $<1$) are still able to better represent the input image. This is because the LMM is able to use the other natural language description from the Prompt 1 output to somewhat successfully recreate the input image, even if the grid configuration and colors are not excplicitly specified.

\subsection{Generative Anomalies in Text-to-Image Generation}
After the LMM has generated output images with Prompt 2 (text-to-image), manual processing of these images was needed to sample their grids and colors (see Section~\ref{sec:processing}).
Our goal in sampling was to rely on clearly defined criteria but also include as many grids as possible, where feasible. Thus, most grids with valid color schemes could be sampled and included in further evaluation.

\begin{figure}[ht]
    \centering

    \begin{subfigure}[t]{0.155\textwidth}
        \centering
        \includegraphics[width=\linewidth]{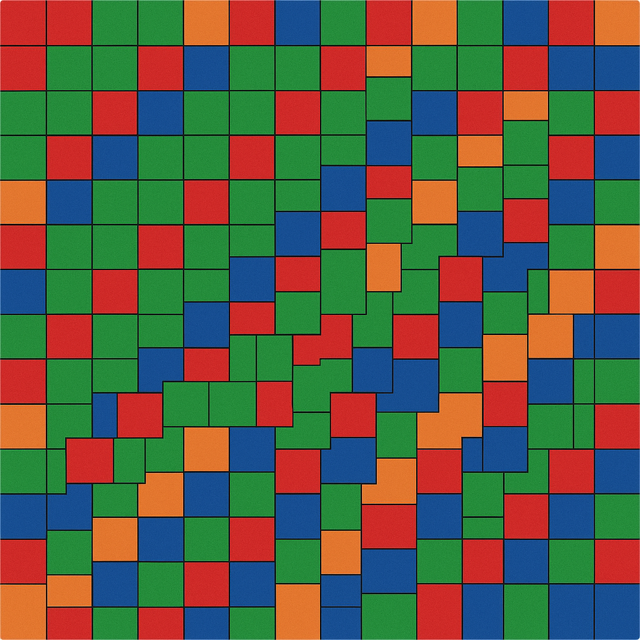}
        \caption{target grid size: $9\times9\times4$; invalid grid with valid colors: different shapes; \textcolor{lightpink}{omitted}.}
        \label{fig:invalid_image_examples_a}
    \end{subfigure}\hfill
    \begin{subfigure}[t]{0.155\textwidth}
        \centering
        \includegraphics[width=\linewidth]{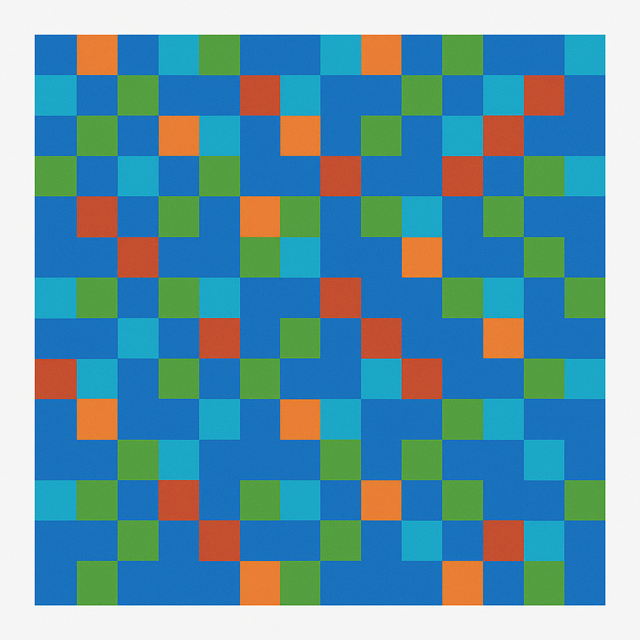}
        \caption{target grid size: $10\times10\times4$; invalid grid with valid colors: no grid lines; \textcolor{lightgreen}{included} (as merged cells).}
        \label{fig:invalid_image_examples_b}
    \end{subfigure}\hfill
        \begin{subfigure}[t]{0.155\textwidth}
        \centering
        \includegraphics[width=\linewidth]{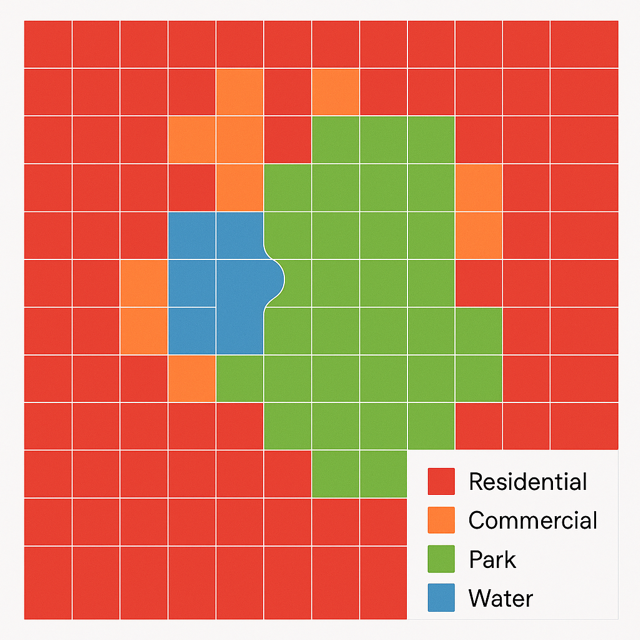}
        \caption{target grid size: $10\times10\times4$; invalid grid with valid colors: merged and covered cells, curved grid line; \textcolor{lightpink}{omitted}.}
        \label{fig:invalid_image_examples_c}
    \end{subfigure}\hfill
    \begin{subfigure}[t]{0.155\textwidth}
        \centering
        \includegraphics[width=\linewidth]{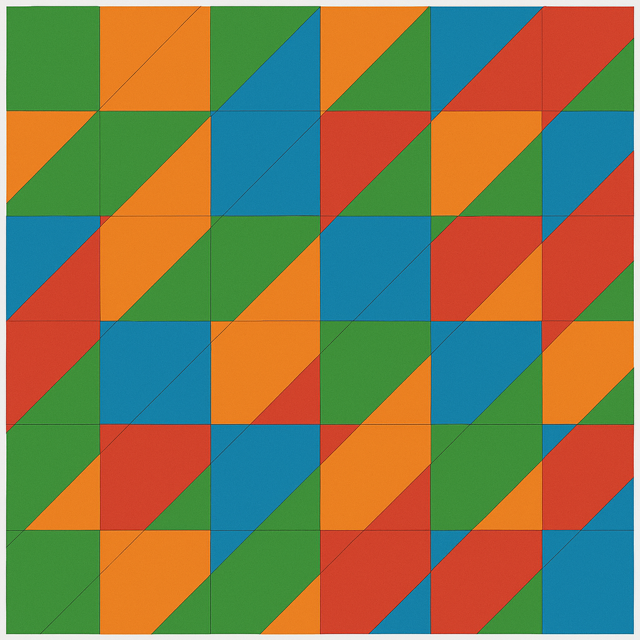}
        \caption{target grid size: $6\times6\times4$; valid grid with invalid colors: diagonal color mix; \textcolor{lightpink}{omitted}.}
        \label{fig:invalid_image_examples_d}
    \end{subfigure}\hfill    
    \begin{subfigure}[t]{0.155\textwidth}
        \centering
        \includegraphics[width=\linewidth]{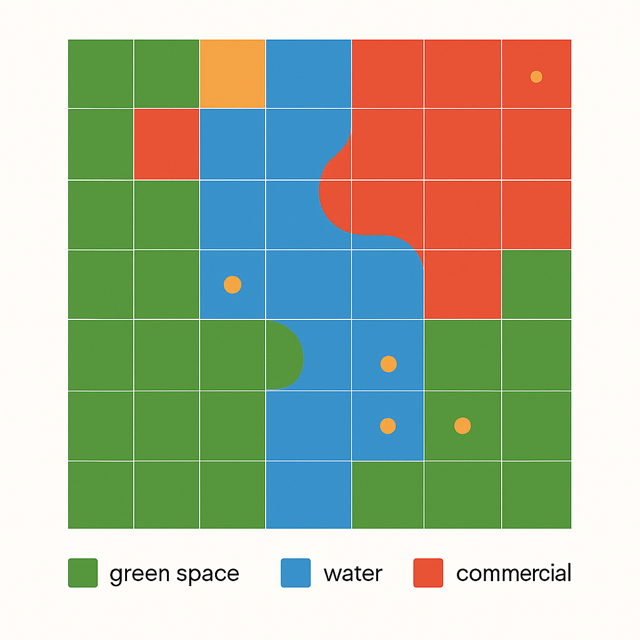}
        \caption{target grid size: $7\times7\times4$; valid grid with invalid colors: river-like structure and dots; \textcolor{lightpink}{omitted}.}
        \label{fig:invalid_image_examples_e}
    \end{subfigure}\hfill
    
    \vspace{7pt}
        
    \begin{subfigure}[t]{0.155\textwidth}
        \centering
        \includegraphics[width=\linewidth]{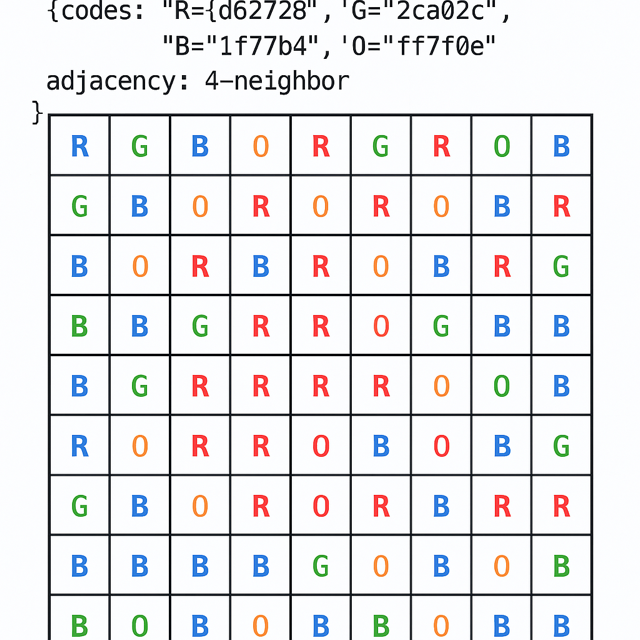}
        \caption{target grid size: $8\times8\times4$; valid grid with invalid colors: empty/white cells; \textcolor{lightgreen}{included}.}
        \label{fig:invalid_image_examples_f}
    \end{subfigure}\hfill
    \begin{subfigure}[t]{0.155\textwidth}
        \centering
        \includegraphics[width=\linewidth]{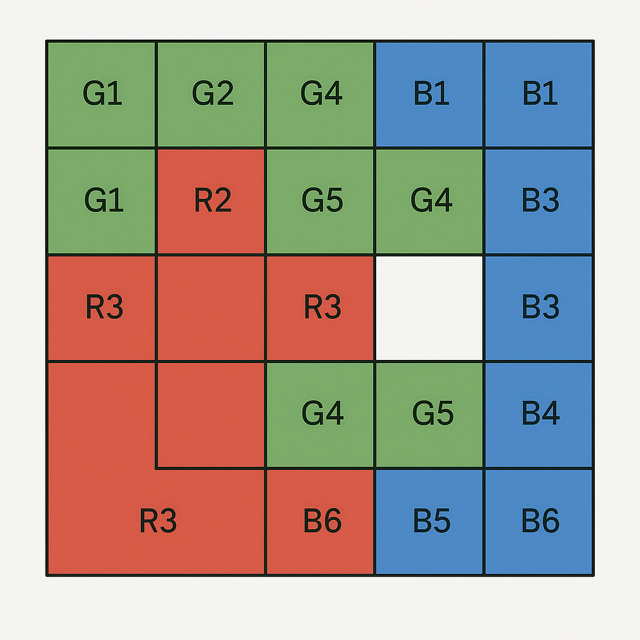}
        \caption{target grid size: $6\times6\times3$; invalid grid and colors: merged and empty/white cells; \textcolor{lightgreen}{included}.}
        \label{fig:invalid_image_examples_g}
    \end{subfigure}\hfill
    \begin{subfigure}[t]{0.155\textwidth}
        \centering
        \includegraphics[width=\linewidth]{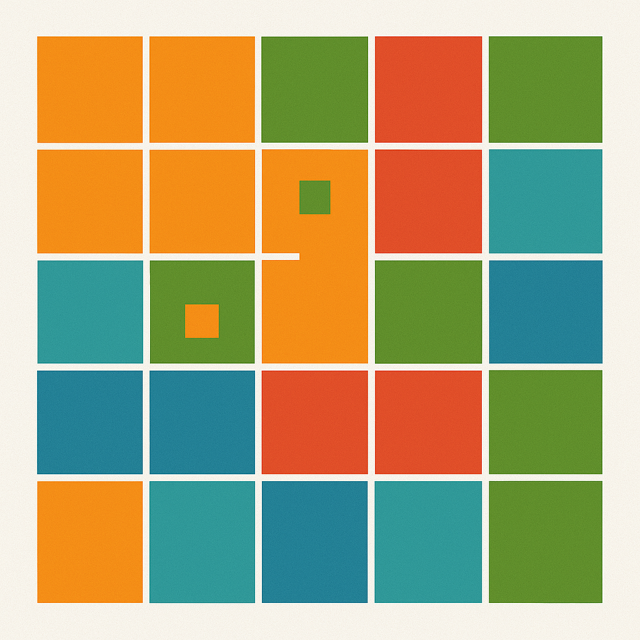}
        \caption{target grid size: $6\times6\times5$; invalid grid and colors: interrupted grid and colored symbols within cells; \textcolor{lightpink}{omitted}.}
        \label{fig:invalid_image_examples_h}
    \end{subfigure}\hfill        
    \begin{subfigure}[t]{0.155\textwidth}
        \centering
        \includegraphics[width=\linewidth]{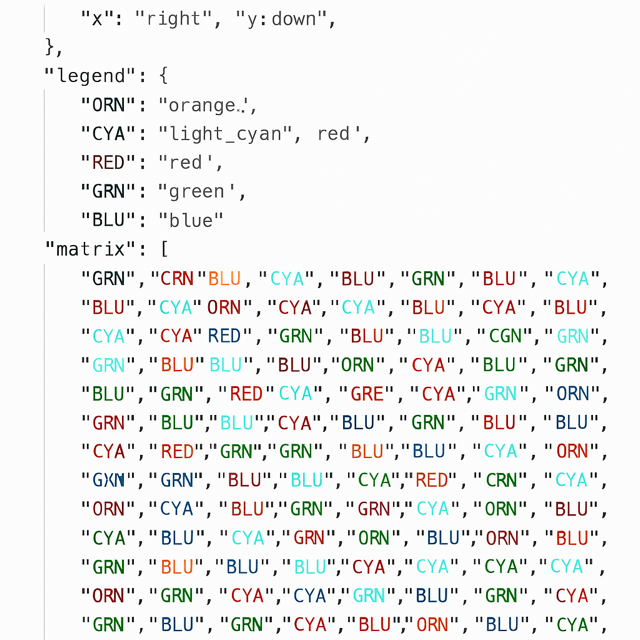}
        \caption{target grid size: $9\times9\times5$; invalid grid and colors: missing grid visualisation; \textcolor{lightpink}{omitted}.}
        \label{fig:invalid_image_examples_i}
    \end{subfigure}\hfill
    \begin{subfigure}[t]{0.155\textwidth}
        \centering
        \includegraphics[width=\linewidth]{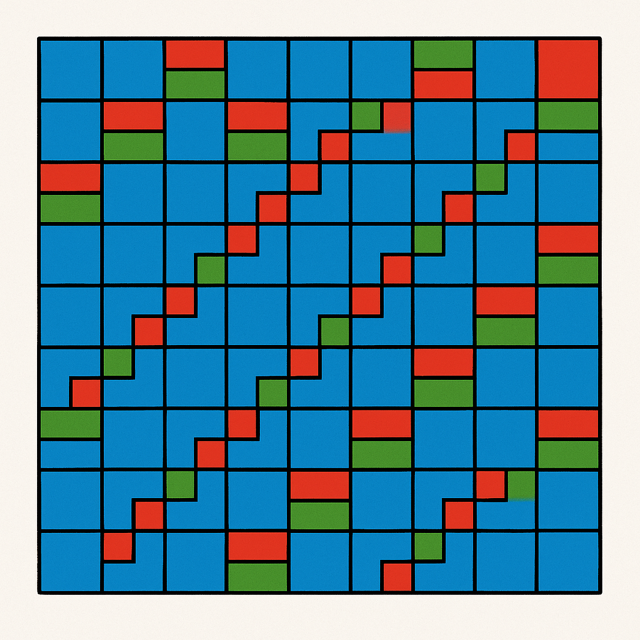}
        \caption{target grid size: $9\times9\times5$; invalid grid and colors: different shapes and color blurring; \textcolor{lightpink}{omitted}.}
        \label{fig:invalid_image_examples_j}
    \end{subfigure}\hfill
        
    \caption{The following examples illustrate the invalid results of text-to-image generation. The target grid sizes, validity, and decision of inclusion or omittance in the process of image sampling (manual color capturing) are indicated below each image.}
    \label{fig:invalid_image_examples}
\end{figure}

In our criteria, we assume that a valid color scheme is given when one clearly defined and closed cell (no matter what shape) is filled with a single color or uniform coloring. Therefore, grids with white or unfilled cells were also taken into account (see Figures~\ref{fig:invalid_image_examples_f} and \ref{fig:invalid_image_examples_g}). This does not affect the evaluation, as such cases appear to be incorrect anyway. Examples where this is not the case (i.e., color scheme is invalid) are shown in Figures~\ref{fig:invalid_image_examples_d}, \ref{fig:invalid_image_examples_e}, \ref{fig:invalid_image_examples_h}, and \ref{fig:invalid_image_examples_j}. Such cases were ommitted from further processing as cell colors could not be sampled and they were assigned maximum errors in the evaluation.

Borderline cases where grids are cut off (see Figure~\ref{fig:invalid_image_examples_f}) have been included, as well as grids with merged cells that follow the lines of the otherwise valid rectangular grid (see Figure~\ref{fig:invalid_image_examples_g}), as long as cells are closed and clearly defined. In contrast to grids with merged cells of other shapes like shown in Figure~\ref{fig:invalid_image_examples_c} or grids with interrupted grid lines as shown in Figure~\ref{fig:invalid_image_examples_h}.
If a grid appears to have no visible lines as shown in Figure~\ref{fig:invalid_image_examples_b} it is a borderline case as well. However, grids that do not have neighbouring cells of equal color and that do not have grid lines are still sampled as valid grids. Nevertheless, the proposed solution is to treat neighbouring cells with the same color as merged cells in order to reflect this in the sampled data. Merged cells are generally sampled once, with the cells being captured in the direction from top left to bottom right at the time of their initial appearance. Grids like shown in Figure~\ref{fig:invalid_image_examples_g} are therefore treated as invalid in the evaluation and receive higher error scores.
However, grids with irregularly shaped cells could not be taken into account within the sampling process, as no clear division into rows and columns is possible here. Figure~\ref{fig:invalid_image_examples_a} shows an example of an invalid grid with valid colors which had to be omitted for further evaluation.

Clearly, grids lacking meaningful semantic geometry, valid grid structures, or consistent coloring could not be encoded into discrete 2D grids. As these severe generative failures represent a complete loss of spatial determinism, they were deemed entirely invalid and had to be omitted from the quantitative analysis (see Figures~\ref{fig:invalid_image_examples_i} and \ref{fig:invalid_image_examples_j}).


\subsection{Diagnostic Analysis of Modality Transfer}
By analysing the LMM performance through the lens of Spatial Error Diagnostic Matrix (Figure \ref{fig:diagnostic_scatter_matrix}, and Table \ref{tab:error_matrix}), we successfully categorize the nature of these breakdowns. In low-complexity scenarios (Figure \ref{fig:diagnostic_scatter_matrix_3_colors}), the grid samples mostly aggregate in the bottom left quadrant (\textbf{Minor Attribute Eorror}, Low Lev. and Low EMD). This establishes an acceptable baseline understanding of spatial arrangement, committing several minor localized color substitutions while preserving the overarching geometric shape.

As the grid and color complexity increases, in a 4-color palette (Figure \ref{fig:diagnostic_scatter_matrix_4_colors}),  the grid samples disperse toward the first (High Lev. and Low EMD) and fourth (High Lev. and High EMD) quadrants, implying a failure of topology and showing hints of model hallucination. In this stage, while the LMMs can grasp the general colors of grid but fail to anchor them into correct, absolute coordinate geometries. The transition to the 5-color palette (Figure \ref{fig:diagnostic_scatter_matrix_5_colors}) shows a descent into catastrophic spatial failure. The dense grid clustering extending into the top-right quadrant (\textbf{Severe Hallucination}, High Lev. and High EMD) indicates a complete collapse of cross-modal translation. In these cases, the models fail to preserve both the 1D sequential order and the 2D geometric boundaries, resulting in images where blocks are either entirely omitted, scattered randomly, or drawn using out-of-bounds categorical colors.

\subsection{Limitations}
While our experimental design establishes a quantifiable baseline for spatial modality transfer, several limitations need to be discussed. Currently, our empirical evaluation is limited to a narrow selection of proprietary flagship models, specifically recent iterations from OpenAI (\texttt{gpt-5} and \texttt{gpt-image-1}). While there is little reason to expect another LMM would suddenly perform this task perfectly, other models need to be tested to gain a wholistic overview of the current state-of-the-art. Additionally, the prompts in this study were executed strictly under zero-shot conditions, leaving the potential performance gains of few-shot prompting or domain-specific fine-tuning unexplored. It is thus possible that a few-shot performance would show significant improvements, but it is questionable whether it would allow for a seamless experience of working with GIS agents if every communication step between the user and the agent has to be repeated several times --- at least when it involves different modalities.

In addition to being tested in a zero-short setting, our prompts have intentionally only vaguely defined the task at hand. This has possibly led to some of the observed issues such as incomplete or incorrectly structured grids. The model would likely perform better with well defined prompts where the valid output format and execution steps are explicitly defined. We believe that future research on agentic GeoAI calls for a systematic investigation of the effects of vague versus well-structured prompts on the agents' success.

Furthermore, although the procedurally generated $N\times N$ categorical grids offer a grounded and robust baseline, they represent a highly abstracted and relatively simplistic spatial reasoning task. This works well to expose the fundamental issue of current LMMs which lose spatial information through modality transfer, but is not a good representation of a full-scale GIS workflow. Because foundation models are trained on real-world data (i.e., photographs as well as textual descriptions of spatial scenes) and not grids of colored squares, they seem to struggle to generalize from context-rich real world vision-reasoning tasks~\cite{cheng2024spatialrgpt} to simpler tasks like the one presented here.  Thus, it remains to be investigated if LMMs would perform better if they were tested on more complex and well contextualized workflows embedded into a full-scale GIS project environment. If the modality transfer was not performed in isolation on simple grid images, but on the real-world land cover maps in a software like ESRI ArcGIS, irregular shapes and additional information about them would make the task more complex, but the added context may also improve the comprehension of the LMM.

Another limitation within our current approach is the GUI-based sampling procedure in the evaluation of the generated images. In its current format, this step is not scalable to larger sets of generated images and remains vulnerable to human errors and subjective interpretation of the operator. For those reasons, it would be better to replace this step with an automated computer vision approach in the future to improve reproducibility and usability. We also only provide the joint evaluation of both modality transfer steps, while it may be informative to consider them in isolation -- i.e., use ground truth input text for the text-to-image prompt. This would allow for a more detailed and objective evaluation of individual modality transfers without allowing the errors in the output of the first step to influence the second modality transfer step.


\section{Conclusion}
\label{sec:conclsuion}

This study focuses on the fundamental issue of LMMs losing geographical information simply due to transferring it from textual to image modality or vice-versa. Anecdotally, this can often be observed when an LMM is asked to generate an image based on a textual description or to textually describe an input image (e.g., a map), and the results returned are imperfect. We argue that a (near) lossless transfer of geographical information between text and image modalities is a pre-requisite for enabling the development of autonomous GIS agents that can adequately perform GIS workflows. 

To this end, we propose an LMM modality transfer task where a set of input images, here regular grids of colored squares, are to be transferred from image to text modality with Prompt 1 and then, based on the textual description outputs of Prompt 1, again be transferred from text to image modality in Prompt 2. This represents a simplified example of working with a land-cover dataset in a GIS workflow. Both outputs of Prompt 1 (generated text) and Prompt 2 (generated image) are then evaluated against the input image to quantify the amount of information lost during modality transfer.

We executed the modality transfer task on two recent OpenAI models (gpt-5 for Prompt 1 and gpt-image-1 for Prompt 2) using 900 input image with increasing grid sizes ($5\times5$ to $10\times10$) and number of colors (3 to 5).
Our results demonstrate that, while LMMs may process vast prior core knowledge learned from large image or language corpus, they still suffer from spatial information loss and hallucination during cross-modal translation, especially on inputs with larger grids, proving that lossless modality transfer remains a task that is yet profoundly difficult for state-of-the-art LMMs.

Future work should first apply this task to a wider array of state-of-the-art LMMs from different producers. It should also consider performing this task in a few-shot manner to avoid the issue where explicit full-grid textual representations are not generated for inputs with larger grid sizes. We also discussed the possibility that tasks using real-world geographic information instead of synthetic examples as input may, even though they seem more complex at first, yield better comprehension in LMMs as they would provide more context and are more similar to the raw training data that the LMMs are trained on.



\bibliography{main}

\appendix

\end{document}